\documentclass[lettersize,journal]{IEEEtran}
\usepackage{amsmath,amsfonts}
\usepackage{algorithmic}
\usepackage{algorithm}
\usepackage{array}
\usepackage{textcomp}
\usepackage{stfloats}
\usepackage{url}
\usepackage{verbatim}
\usepackage{graphicx}
\usepackage{cite}
\usepackage[T1]{fontenc}
\usepackage{colortbl}
\usepackage{subfigure}
\usepackage{tabularx,booktabs}
\usepackage{enumitem}
\usepackage{amssymb}
\usepackage{multirow}
\usepackage{pifont}
\usepackage{todonotes}
\usepackage{hyperref}
\hypersetup{colorlinks,linkcolor={blue},citecolor={blue},urlcolor={black}} 
\usepackage{breakcites}

\newcommand{\xmark}{\ding{55}}%
\newcommand{\cmark}{\ding{51}}%
\newcommand{\ouralg}{KIF} 
\newcommand{\ouralgn}{KIFLoRA} 
\newcommand{\ouralgm}{KIF-M} 

\begin{document}

\title{
KIF: Knowledge Identification and Fusion for Language Model Continual Learning 
}

\author{Yujie Feng, Xu Chu, Yongxin Xu, Zexin Lu, Bo Liu, Philip S. Yu and Xiao-Ming Wu
\thanks{Yujie Feng, Zexin Lu, Bo Liu, and Xiao-Ming Wu are with the Department of Computing, The Hong Kong Polytechnic University, Hong Kong S.A.R., China. E-mail: \{yujie.feng, zexin.lu, bo.liu\}@connect.polyu.hk, xiao-ming.wu@polyu.edu.hk}
\thanks{Xu Chu and Yongxin Xu are with the School of Computer Science, Peking University, Beijing, China. E-mail: chu\_xu@pku.edu.cn, xuyx@stu.pku.edu.cn}
\thanks{Philip S. Yu is with the Department of Computer Science, University of Illinois at Chicago, USA. E-mail: psyu@uic.edu}
\thanks{This work is an extension of our ACL 2024 paper, available at \url{https://aclanthology.org/2024.acl-long.69.pdf}.}
\thanks{Our code is available at \url{https://github.com/WoodScene/TaSL}.}}

\markboth{Journal of \LaTeX\ Class Files,~Vol.~14, No.~8, August~2021}%
{Shell \MakeLowercase{\textit{et al.}}: A Sample Article Using IEEEtran.cls for IEEE Journals}


\maketitle

\begin{abstract}
Language model continual learning (CL) has recently attracted significant interest for its ability to adapt large language models (LLMs) to dynamic real-world scenarios without retraining. A major challenge in this domain is catastrophic forgetting, where models lose previously acquired knowledge upon learning new tasks. Existing approaches commonly utilize multiple parameter-efficient fine-tuning (PEFT) blocks to acquire task-specific knowledge, yet these methods are inefficient and fail to leverage potential knowledge transfer across tasks.
In this paper, we introduce a novel CL framework for language models, named \underline{K}nowledge \underline{I}identification and \underline{F}usion (KIF), which boosts knowledge transfer without depending on memory replay. {\ouralg} initially segregates the model into `skill units' based on parameter dependencies, allowing for more precise control. Subsequently, it employs a novel group-wise knowledge identification technique to ascertain the importance distribution of skill units for a new task. By comparing this importance distribution with those from previous tasks, we implement a fine-grained knowledge fusion strategy that retains task-specific knowledge, thereby preventing forgetting, and updates task-shared knowledge, which facilitates bi-directional knowledge transfer.
As a result, {\ouralg} achieves an optimal balance between retaining prior knowledge and excelling in new tasks.
{\ouralg} also demonstrates strong generalizability, making it suitable for various base models and adaptable to PEFT methods like LoRA. Furthermore, it offers notable extensibility, supporting enhancements through integration with memory replay techniques.
Comprehensive experiments conducted on two CL benchmarks, involving models ranging from 220M to 7B parameters, affirm the effectiveness of {\ouralg} and its variants across different settings.

\end{abstract}


\section{Introduction}
\IEEEPARstart{E}{quipping} large language models (LLMs) with continual learning (CL) capabilities to sequentially learn different tasks is essential for their deployment in real-world scenarios~\cite{brown2020language, de2021continual}. This capability enables LLMs to dynamically adapt to new tasks and acquire additional knowledge~\cite{wang2024comprehensive, long2018transferable}.
An effective CL system must address two critical challenges:
(1) Catastrophic Forgetting (CF) ~\cite{mccloskey1989catastrophic}, where a model's performance on previous tasks deteriorates as it learns new ones,
and (2) Knowledge Transfer (KT) ~\cite{ke2021achieving}, enhancing task performance through the transfer of knowledge. 
KT can be categorized into forward transfer, which improves new task performance using knowledge from previous tasks, and backward transfer, which enhances performance on previous tasks after learning a new relevant task. 
Achieving a balance between retaining previous knowledge and excelling in new tasks is vital for success.

\begin{figure}[t]
  \centering
  \includegraphics[width=1.0\linewidth]{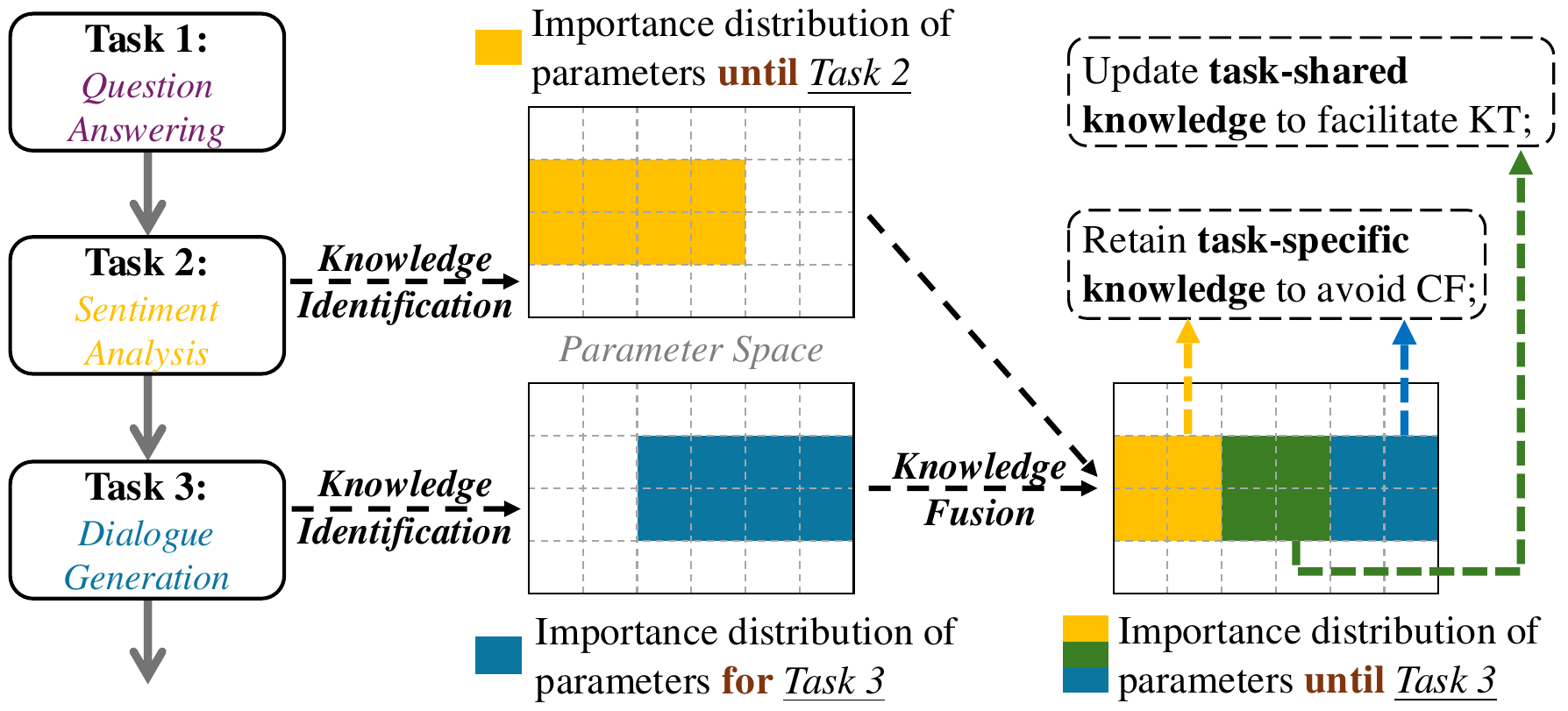}
  \caption{
  Conceptual illustration of {\ouralg}. By identifying task-relevant areas across both previously accumulated and current tasks, we can consolidate the task-shared and task-specific parameters to facilitate efficient knowledge transfer (KT) and mitigate catastrophic forgetting (CF).
  }
  \label{fig:intro}
\end{figure}

Due to the high computational demands, recent studies have explored CL for LLMs using parameter-efficient fine-tuning (PEFT) methods~\cite{hu2021lora, ding2022delta}, such as Low-Rank Adaptation (LoRA) \cite{hu2021lora}.
Traditional rehearsal-based approaches, which require storing data examples from past tasks for replay, face significant privacy and memory issues~\cite{shu2023omni, wang2024comprehensive}.

Another prominent line of work in recent studies is the parameter isolation CL methods~\cite{rusu2016progressive, wang2023rehearsal, jin2024one, wang2024unified}. 
These methods assign a dedicated PEFT block for each new task to capture task-specific knowledge, preserving the block for selective activation during testing. However, they exhibit significant limitations in effectively addressing the KT and CF challenges.

\textbf{On one hand}, the prevalent design of these frameworks, characterized by the use of multiple PEFT blocks, introduces redundancy and operational inefficiency \cite{zhu2024model}. As the number of tasks grows, the required PEFT blocks accumulate, leading to increased memory storage demands and making it difficult to handle long task sequences \cite{zhao2024sapt, liang2024inflora}. Furthermore, such approaches tend to develop separate specialist models for each task, which severely limits their capability to generalize across unseen tasks \cite{razdaibiedina2022progressive} (\textit{\textbf{Limitation 1}}). 

\textbf{On the other hand}, these parameter-isolation CL methods overlook the interactions between tasks, which hinders KT. 
For instance, the method in \cite{wang2023rehearsal} learns each PEFT block separately within individual tasks. Similarly, Orthogonal Low-Rank Adaptation (O-LoRA) \cite{wang2023orthogonal} employs orthogonal subspace gradient projections for parameter updates.
While these strategies might mitigate CF, they cut off the potential for leveraging knowledge distributed across various PEFT blocks, thus obstructing bi-directional KT among tasks and resulting in less-than-optimal performance (\textit{\textbf{Limitation 2}}).

To address these limitations, we introduce \underline{\textbf{K}}nowledge \underline{\textbf{I}}dentification and \underline{\textbf{F}}usion ({\ouralg}) \cite{feng2024tasl}, a novel CL framework designed to improve KT between tasks without relying on memory replay.
Our approach is motivated by recent findings that model parameters contribute unevenly to performance~\cite{panigrahi2023task}. For instance, the authors in \cite{zhang2024unveiling} uncovered a core region in LLMs that is crucial for all languages, suggesting that preserving these vital regions can mitigate forgetting.
Additionally, findings from \cite{zhu2024model} indicate that fine-tuning with LoRA often preserves many unnecessary parameter changes, pointing to substantial redundancies within PEFT blocks.

Based on these insights, {\ouralg} facilitates KT by pinpointing and consolidating the importance distribution of model parameters across tasks.
Initially, {\ouralg} utilizes a group-wise importance-aware \textit{\textbf{knowledge identification}} technique that employs gradient trajectories to identify tiny regions within the parameter space that store crucial knowledge for the current task. By comparing the importance distribution with those of previous tasks, we can then differentiate between task-specific and task-shared regions, as illustrated in Figure \ref{fig:intro}.
The subsequent \textit{\textbf{knowledge fusion}} phase then categorically integrates weights from previous tasks with the current one, efficiently mitigating CF.

In detail, {\ouralg} first reconstructs the model or PEFT block into fine-grained ``\textit{\textbf{skill units}}''. A skill unit refers to a distinct subset of model parameters that encapsulates specific functional capabilities or knowledge relevant to a particular task, such as the Query matrix within the self-attention layer. By operating at this finer granularity, we can localize and consolidate task-specific and shared knowledge within a single PEFT block, rather than adapting a separate PEFT block for each task as in previous works (\textbf{addressing Limitation 1}).

The importance-aware knowledge identification method employs a new group-wise metric to compute importance scores, effectively quantifying the significance of each skill unit for the current task. 
Our knowledge fusion phase, then based on a fine-grained model averaging strategy, effectively manages different types of knowledge.
This stage facilitates forward KT by using previously fine-tuned weights as the starting point for new tasks. For backward KT, we merge knowledge from both current and past tasks into localized task-shared skill units, boosting their effectiveness. To prevent CF, we ensure the integrity of skill units containing prior task-specific knowledge is maintained, safeguarding them from being altered by the learning of new tasks (\textbf{addressing Limitation 2}).


Given the widespread adoption and success of LoRA in fine-tuning LLMs, there is a compelling opportunity to optimize the {\ouralg} framework specifically for LoRA. 
While the initial {\ouralg} framework has shown promising results, its validation has been limited to a specific sub-task, and its broader applicability on general CL benchmarks remains explored. Additionally, the reliance on first-order gradients for knowledge identification in {\ouralg} may not yield precise importance localization. The knowledge fusion phase also involves many hyperparameters, which could complicate the averaging process.

To address these issues, we propose {\ouralgn}, a variant of {\ouralg} optimized for LoRA. 
{\ouralgn} redesigns the LoRA adapter into new skill units based on parameter dependencies, streamlining knowledge management throughout sequential task learning. It incorporates an orthogonal loss during fine-tuning to ensure distinct latent semantic features are captured within the same LoRA adapter. For knowledge identification, {\ouralgn} utilizes a new second-order gradient approximate group-wise metric, enhancing the precision of importance scores.
In knowledge fusion, it shifts from hard-mask to soft-mask averaging, employing an adaptive technique that flexibly integrates task-specific and shared parameters according to the importance of the skill unit.
Additionally, we combine {\ouralg} with memory replay to broaden its application scope, introducing the {\ouralgm} model. {\ouralg} framework and its variants excel in mitigating CF and showcase remarkable capabilities for KT, outperforming SOTA methods.

The main contributions are summarized as follows:
\begin{itemize}
\item 
We introduce the Knowledge Identification and Fusion ({\ouralg}) \textbf{framework} for language model continual learning. By pinpointing and amalgamating task-specific and task-shared knowledge at a granular skill unit level, {\ouralg} ensures effective knowledge transfer and significantly reduces catastrophic forgetting, addressing the shortcomings of previous methodologies.

\item 
We devise various group-wise knowledge identification and fine-grained knowledge fusion \textbf{techniques}. Notably, our {\ouralg} utilizes either first-order gradients or an innovative second-order gradient approximation for assessing parameter importance. For knowledge fusion, we implement strategies ranging from categorical hard-masking to adaptive soft-masking, enhancing the flexibility and precision of knowledge integration.

\item 

The {\ouralg} framework is distinguished by its robust \textbf{generalizability} and \textbf{extensibility}. 
Its skill units are designed flexibly, allowing for seamless tailoring to PEFT methods, such as LoRA, optimizing performance across different model architectures. Moreover, integrating {\ouralg} with memory replay enables further performance enhancements and broadens its applicability to diverse scenarios.

\item 


Extensive \textbf{evaluation} on two CL benchmarks demonstrates the superiority of our {\ouralg} framework and its variants in facilitating knowledge transfer and mitigating catastrophic forgetting, especially in memory-free scenarios. 
Furthermore, {\ouralg} consistently excels across diverse model sizes (from 220M to 7B), various model architectures (T5 and LLaMA-2), and unseen tasks.

\end{itemize}

\section{Related Work}

\subsection{Language Model Continual Learning}
Continual learning \cite{wang2024comprehensive} seeks to develop algorithms that accumulate knowledge from non-stationary data sources.

\textbf{Conventional continual learning} methods can be categorized into three main types:
(i) Regularization-based methods impose explicit constraints to preserve the knowledge of previous tasks by penalizing the variation of each parameter based on its contribution to past tasks~\cite{jin2024one, li2023crnet}, such as in the EWC \cite{kirkpatrick2017overcoming} method. 
However, these methods typically restrict the learning of task-shared knowledge by only considering parameter importance from the perspective of past tasks. This limitation often leads to a suboptimal balance between retaining previous knowledge and excelling at new tasks, as it does not account for the relevance of parameters to both past and current tasks.
(ii) Rehearsal-based methods mitigate catastrophic forgetting by either storing old training samples \cite{wang2024inscl, pham2023continual} or training generative models to provide pseudo-samples of previous tasks \cite{huang2024mitigating, li2023variational}, enabling data replay during the learning of new tasks.
(iii) Architecture-based methods aim to prevent task interference by dynamically expanding model capacity or assigning dedicated parameter subspaces to each task \cite{rypesc2024divide, xu2021adaptive}. While effective at minimizing forgetting, these methods often hinder knowledge transfer (KT) between tasks.
Our {\ouralg} framework stands out by bi-directionally analyzing parameter importance across both historical and current tasks, enabling effective KT while mitigating forgetting. To further enhance performance, we extend {\ouralg} with memory replay techniques, resulting in {\ouralgm}, improving its flexibility and applicability across various scenarios.

\textbf{Language model continual learning with PEFT} methods in the era of LLMs often adopt a parameter isolation strategy. This approach involves using a pipeline to learn and select PEFT blocks for each task \cite{wang2023rehearsal, chen2023towards}. However, as the number of tasks increases, the accumulation of PEFT blocks can become inefficient, and achieving KT between blocks remains challenging.
More recently, authors in \cite{li2024theory} have explored using the Mixture of Experts (MoE) approach to connect multiple PEFT blocks, but this essentially relies on coarse-grained model averaging. This can lead to overemphasizing unimportant weights, contaminating previously acquired task-specific knowledge, and causing forgetting.
To address these challenges in the context of PEFT-based CL, we extend the {\ouralg} framework and introduce {\ouralgn}. By dividing fine-grained skill units based on parameter dependencies within LoRA, {\ouralgn} enhances KT efficiency, enables more precise control, and effectively mitigates forgetting.



\subsection{Knowledge Identification}
Research has shown that model parameters contribute unevenly to performance \cite{michel2019sixteen}. For example, the authors in \cite{zhang2024unveiling} identified a core region in LLMs linked to linguistic competence; freezing this region during further pre-training has been shown to mitigate catastrophic forgetting. Additionally, the authors in \cite{chen2024learnable} found neurons within LLMs that store personally identifiable information, highlighting the potential for improved privacy protections. However, these studies primarily focus on describing such findings rather than applying them to solve practical problems.

Motivated by these findings, we address the challenge of catastrophic forgetting in CL. While \cite{panigrahi2023task} introduced the concept of ``skill localization'' to identify critical parameters in pre-trained language models, their approach requires additional steps to identify and retrain these parameters post-fine-tuning, which impacts efficiency.
In contrast, our importance-aware knowledge identification method leverages trajectory gradients during each task's training phase to identify parameter importance, significantly reducing processing time. Combined with our knowledge fusion strategy, which explicitly targets task-specific and shared parameters, our approach effectively mitigates forgetting and achieves knowledge transfer in CL.




\section{Preliminaries}
\subsection{Continual Learning Setup}

Continual learning~\cite{de2021continual} aims to develop algorithms that can progressively accumulate knowledge from ongoing sequences. In supervised continual learning, a series of tasks $\{\mathcal{T}_1, \ldots, \mathcal{T}_K \}$ is presented sequentially. Each task $\mathcal{T}_k$ includes a distinct target dataset $\mathcal{D}_k = \left\{ \left( x_i^k, y_i^k \right) \right\}_{i=1}^{N_k}$ of size $N_k$, where $x_i^k \in \mathcal{X}_k$ and $y_i^k \in \mathcal{Y}_k$. 
The model, parameterized by $\Theta$, is trained to adapt to these tasks one at a time, accessing only $\mathcal{D}_k$ during the $k$-th task. The overarching goal of continual learning is to optimize the following objective:

\begin{equation}
\max_{\Theta} \sum_{k=1}^{K} \sum_{x,y \in \mathcal{D}_k} \log p_{\Theta}(y \mid x)
\end{equation}

And we denote $f_k$ as the model trained up to task $\mathcal{T}_k$, while $\hat{f}_{k}$ represents the model after integrating knowledge from $\hat{f}_{k-1}$ and $f_k$ through averaging.
Our {\ouralg} framework is designed to tackle a more challenging scenario where the model cannot access any historical data during training \cite{wang2023orthogonal}.
To broaden its applicability, we have augmented {\ouralg} with memory replay. In this enhanced setup, known as {\ouralgm}, we store a random subset of $\left| \mathcal{M} \right|$ samples from each prior task's training set $\mathcal{D}_i$ in memory $\mathcal{M}_i$. The model is then trained jointly on the new task data $\mathcal{D}_k$ and the accumulated memory $\mathcal{M}_{<k}$, leveraging this memory replay to improve performance.

\subsection{Low-Rank Adaption}

Low-Rank Adaption (LoRA)~\cite{hu2021lora} is a parameter-efficient fine-tuning method to adapt LLMs to novel tasks.
It operates under the assumption that parameter changes during full fine-tuning on a downstream task reside within a low-rank space. This approach allows for not having to fine-tune the entire model, thus significantly improving computational efficiency and resource utilization.
Inspired by the low-rank internal dimensionality~\cite{aghajanyan2020intrinsic},  LoRA hypothesizes the updates to the weights also has a low ``intrinsic rank'' during adaptation.
For instance, consider a pre-trained weight matrix $W^{(0)} \in \mathbb{R}^{out \times in}$ that takes $x$ as input, LoRA modifies the output $h$ of $W^{(0)}$ with a low-rank decomposition:
\begin{equation}
h = W^{(0)} x + \Delta W x = W^{(0)} x + B A x,
\label{eq:lora}
\end{equation}
while $B \in \mathbb{R}^{out \times r}$, $A \in \mathbb{R}^{r \times in}$, and the rank $r \ll \min(in, out)$. $A$ is initialized from a random Gaussian distribution and $B$ is initialized with all zeros.
During training, $W^{(0)}$ remains fixed, and only $B A$ is updated. 
Due to its effectiveness, LoRA has emerged as one of the most favored PEFT methods within the NLP community \cite{qin2024large}.

\section{Proposed Method: KIF}
\subsection{Overview}
\begin{figure*}[t]
  \centering
  \includegraphics[width=0.9\linewidth]{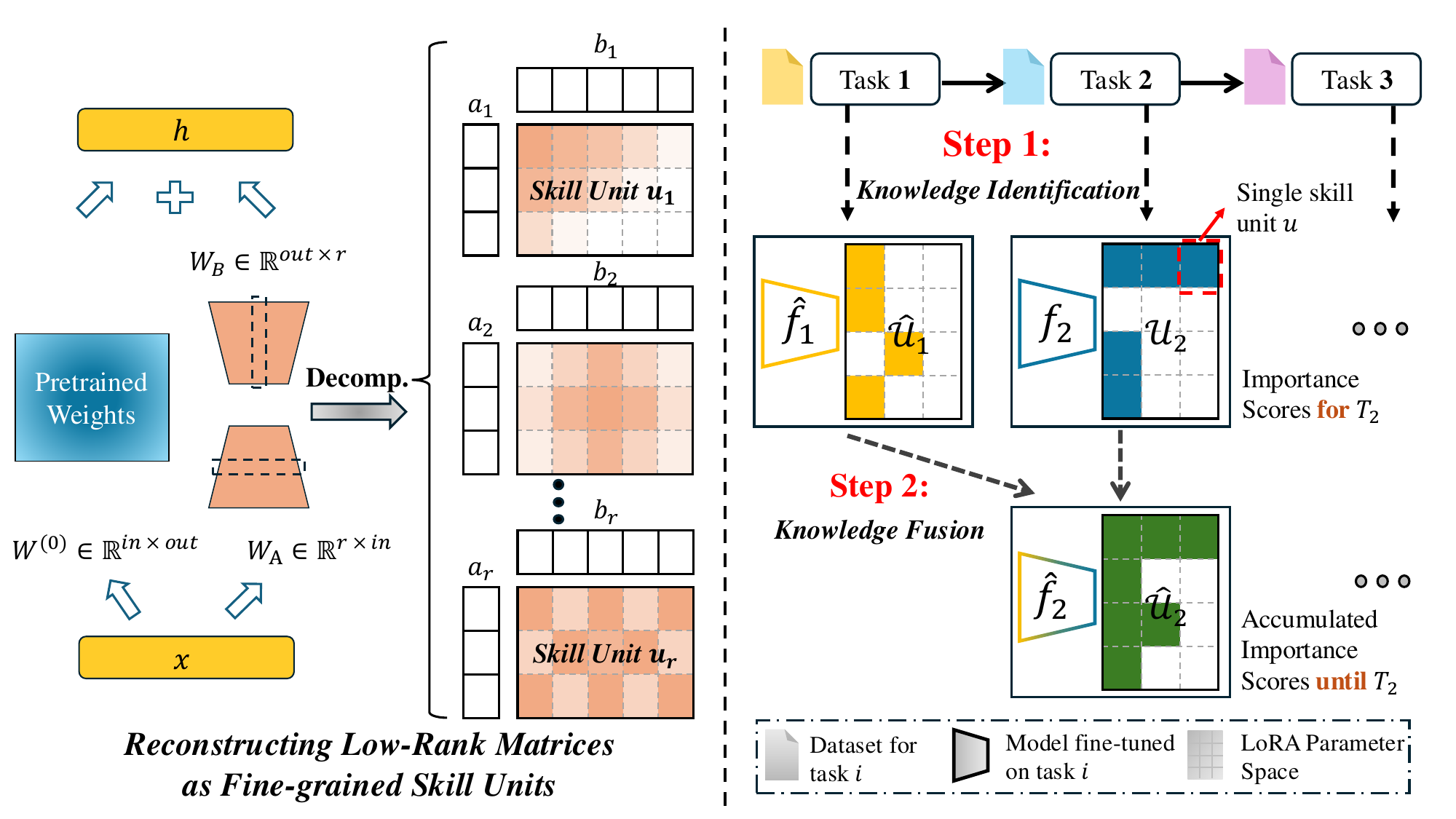}
  \caption{\textit{Left:} Depiction of reconstructing LoRA as fine-grained skill units. \textit{Right:} \textbf{Overview of {\ouralg}.}
\textbf{Step 1}: We compute the importance scores of skill units for the current task $k$ using our importance-aware knowledge identification method during fine-tuning.
\textbf{Step 2}: Based on a categorical model averaging strategy, the knowledge fusion mechanism merges the model $\hat{f}_{k-1}$, which accumulates knowledge of all previous tasks, with the current task's model $f_k$. 
This integration is strategically guided by the importance distributions of skill units across different tasks. This iterative process continues with the addition of each new task.
  }
  \label{fig:method}
\end{figure*}

{\ouralg} includes two key components: (i) \textit{\textbf{Knowledge Identification}}, utilizing a group-wise importance metric to precisely identify the importance distribution of parameters across tasks, and (ii) \textit{\textbf{Knowledge Fusion}}, which employs an innovative fine-grained model averaging strategy to integrate model weights from both current and previous tasks for effective knowledge transfer. 
Figure \ref{fig:method} presents a detailed overview of our proposed {\ouralg} framework, with the following subsections elaborating on each component and the corresponding extensions.

\subsection{Fine-Grained Skill Unit}
Before localizing the importance of parameters, it is essential to define a structure that facilitates better organization and understanding of the knowledge stored within the model. This structure aids in addressing CF and enhancing KT. We define this basic structural element of stored skills or knowledge in the model as a ``skill unit.'' Depending on different usage scenarios, we propose two structures for skill units:

\subsubsection{Matrix-Level Skill Unit}
In this division strategy, we define skill units as individual matrices in the model, such as the query or key matrices in the self-attention layer or the A matrix in LoRA.
This approach aligns with the original {\ouralg} framework and offers the advantage of being model-agnostic. It is suitable for both traditional full-parameter fine-tuning and modern parameter-efficient fine-tuning methods like LoRA, ensuring broad generalizability.

\subsubsection{LoRA-Tailored Skill Unit}
While the matrix-level division is straightforward and intuitive, it does not adequately address intra-matrix redundancy or inter-matrix dependencies.

To address these limitations and the inefficiencies noted in recent work \cite{wang2023orthogonal}, which treats each LoRA adapter as a container for task-specific knowledge, we propose a more refined decomposition of model matrices into new, finer-grained ``skill units'' based on parameter dependencies specifically tailored for LoRA.
The construction process for these LoRA-tailored skill units unfolds as follows.


As shown in Eq (\ref{eq:lora}), $A$ and $B$ can be viewed as a combination of a set of vectors: $A = [a_1, a_2, \cdots, a_r]$, $B = [b_1, b_2, \cdots, b_r]$, where $a_i \in \mathbb{R}^{in}$, $b_i \in \mathbb{R}^{out}$. Thus $B A$ can be further disassembled as:
\begin{equation}
\begin{aligned}
W  &= W^{(0)} + b_1 a_1 + b_2 a_2 + \cdots + b_r a_r \\
&= W^{(0)} + u_1 + u_2 + \cdots + u_r,
\end{aligned}
\end{equation}
where each skill unit $u_i$ is a matrix formed by the product of the vectors $b_i, a_i$.
Thus, LoRA can be conceptualized as an aggregation of knowledge across multiple skill units:
\begin{equation}
\begin{aligned}
W &= W^{(0)} + \sum_{i=1}^{r} b_i a_i = W^{(0)} + \sum_{i=1}^{r} u_i
\end{aligned}
\end{equation}

This division strategy significantly reduces the redundancy inherent in the traditional approach of utilizing separate LoRA adapters for each task \cite{wang2023orthogonal}. To ensure that different skill units within the same layer learn distinct latent semantic features, we incorporate a regularization term as introduced in \cite{zhang2023adaptive}:
\begin{equation}
R(A, B) = \|A^T A - I\|_F^2 + \|B^T B - I\|_F^2,
\end{equation}
where $I$ is the identity matrix, this forces $A$ and $B$ to be orthogonal after training.

\subsection{Importance-aware Knowledge Identification}
To calculate the importance of each skill unit $u$, we introduce a group-wise metric that mitigates the significant computational and storage burdens associated with previous parameter-level importance calculation methods \cite{konishi2023spg}:
\begin{equation}
\mathcal{I}(u) = \frac{1}{in \times out} \sum\limits_{i=1}^{in} \sum\limits_{j=1}^{out} s(w_{ij}) \label{eq:3}
\end{equation}
where $w_{ij}$ denotes the trainable parameters, and $in \times out$ represents the total parameter count in a skill unit $u$. The function $\mathcal{I}(u)$ measures the collective importance of all parameters within each skill unit, with higher values indicating greater importance.
The importance function $s(\cdot)$ for individual parameters, inspired by the pruning community \cite{lecun1989optimal}, is quantified by measuring the impact of its removal on the loss. Specifically, to estimate the importance of $W_{i}$, we measure the change in loss when the parameter is zeroed out using:
\begin{equation}
\begin{aligned}
I_{W_i} = |\Delta \mathcal{L}(\mathcal{D})| = |\mathcal{L}_{W_{i}}(\mathcal{D}) - \mathcal{L}_{W_{i}=0}(\mathcal{D})| \\= \left| \frac{\partial \mathcal{L}(\mathcal{D})}{\partial W_{i}} W_{i} - \frac{1}{2} W_{i} H_{ii} W_{i} + \mathcal{O}(\|W_{i}\|^3) \right| \label{eq:loss}
\end{aligned}
\end{equation}
where $H$ is the Hessian matrix, $W_i$ is the $i$-th parameter in $W$, and $\mathcal{L}$ is the next-token prediction loss. 
If removing a parameter has a significant influence, then the model is sensitive to it, and we should retain it.
Based on Eq. (\ref{eq:loss}), we can derive the following two metrics for calculating importance:

\subsubsection{First-Order Gradient-Based Metric}
In the original {\ouralg} framework, we defined the importance function $s(\cdot)$ as the magnitude of the gradient-weight product:
\begin{equation}
I_{W_{i}}=\left|W_{i} \nabla_{W_{i}} \mathcal{L}\right| \label{eq:1}
\end{equation}

This metric primarily considers the first-order gradient term in determining parameter importance. However, the significance of a parameter is also influenced by second-order gradients, represented by the Hessian matrix. Overlooking this second-order gradient information can introduce biases in the identification process, potentially degrading model performance. Therefore, we propose the following new metric.

\subsubsection{Second-Order Gradient Approximation Metric}
The direct computation of the Hessian matrix for LLMs is impractical due to its $\mathcal{O}(N^2)$ complexity. To address this, we propose a second-order gradient approximation metric that balances the efficiency of first-order methods with the precision of second-order gradients. To reduce computational demands, we approximate the diagonal of the Hessian, $H_{ii}$, using the Fisher information matrix \cite{rissanen1996fisher}. The importance is then defined as:
\begin{equation}
I_{W_{i}} \approx \left| \frac{\partial \mathcal{L}(\mathcal{D})}{\partial W_{i}} W_{i} - \frac{1}{2} \sum_{j=1}^N \left(\frac{\partial \mathcal{L}(\mathcal{D}_j)}{\partial W_{i}} W_{i} \right)^2 \right| \label{eq:ipt}
\end{equation}

However, calculating the importance as specified in Eq. (\ref{eq:ipt}) across the entire training set introduces significant challenges, as model training typically accesses only mini-batch data. This limitation subjects the metric to variability due to stochastic sampling and training dynamics, introducing substantial uncertainty in estimating parameter sensitivity.
To mitigate this, we propose a more reliable importance metric based on sensitivity smoothing and uncertainty quantification \cite{zhang2023adaptive}:
\begin{equation}
\begin{split}
\bar{I}^{(t)}\left(w_{i j}\right)  =\alpha_{1} \bar{I}^{(t-1)}\left(w_{i j}\right)+ \left(1-\alpha_{1}\right) I^{(t)}\left(w_{i j}\right) \label{eq:I}
\end{split}
\end{equation}
\begin{equation}
\begin{split}
\bar{U}^{(t)}\left(w_{i j}\right)  =\alpha_{2} \bar{U}^{(t-1)}\left(w_{i j}\right)+ \\ \left(1-\alpha_{2}\right)\left|I^{(t)}\left(w_{i j}\right)-\bar{I}^{(t)}\left(w_{i j}\right)\right| \label{eq:U}
\end{split}
\end{equation}
where $\alpha_{1}$ and $\alpha_{2}$ are smoothing factors, and $t$ is the iteration number.
$\bar{I}^{(t)}$ represents smoothed sensitivity and $\bar{U}^{(t)}$ is the uncertainty term quantified by the local variation between $I^{(t)}$ and $\bar{I}^{(t)}$.
Using the exponential moving average of the importance metric, we can retain and examine the trajectory gradient for a longer time, yielding a more precise importance assessment.
Importance is ultimately determined by:
\begin{equation}
s^{(t)}\left(w_{i j}\right)=\bar{I}^{(t)}\left(w_{i j}\right) \cdot \bar{U}^{(t)}\left(w_{i j}\right) \label{eq:2}
\end{equation}

To calculate the importance score of each skill unit for the current task $\mathcal{T}_k$, we use Eq. (\ref{eq:3}) during fine-tuning. The model $f$ with $n$ skill units is denoted as $\mathcal{U} = \left\{u_1, \ldots, u_n\right\}$, with their importance scores for task $\mathcal{T}_k$ denoted by $\mathcal{I}(\mathcal{U}_k) \in \mathbb{R}^n$.
The detailed computation process is provided in Algorithm~\ref{alg:ipt}.

\begin{algorithm}[t]
\caption{Importance-aware Knowledge Identification}\label{alg:ipt}
\begin{algorithmic}[0.9]
\renewcommand{\algorithmicrequire}{\textbf{Input:}}
\renewcommand{\algorithmicensure}{\textbf{Output:}}
\REQUIRE Training dataset $\mathcal{D}_k$ for task $\mathcal{T}_k$; total training iterations $T$; hyperparameters $\alpha_1, \alpha_2$.
\FOR {$t = 1, \ldots, T$}
    \STATE Sample a mini-batch from $\mathcal{D}_k$ and compute the gradient $\nabla \mathcal{L}$;
    \STATE Compute the sensitivity $I\left(w_{i j}\right)$ via Eq. (\ref{eq:1}) or Eq. (\ref{eq:ipt});
    \STATE Update $\bar{I}^{(t)}$ via Eq. (\ref{eq:I}) and $\bar{U}^{(t)}$ via Eq. (\ref{eq:U});
\ENDFOR
\STATE Compute the importance score $\mathcal{I}(u^k_i)$ for each skill unit $u^k_i$ by Eq. (\ref{eq:3}), for $i = 1, \ldots, n$.
\ENSURE  $f_k$ and importance scores $\mathcal{I}(\mathcal{U}_k)$ for $\mathcal{U}_k$.
\end{algorithmic} 
\end{algorithm}

After computing the importance scores for each skill unit at the current task $\mathcal{T}_k$, it is crucial to compare these with scores from all previously learned tasks to distinguish between task-specific and task-shared skill units. 
To streamline this process and avoid the inefficiencies associated with storing scores for each past task, we aggregate importance scores from all prior tasks into a cumulative score for tasks up to $\mathcal{T}_{k-1}$. This approach allows for iterative refinement of accumulated scores without the need to separately store past task scores. The skill units with these cumulative scores up to $\mathcal{T}_{k-1}$ are denoted as $\hat{\mathcal{U}}_{k-1}$, calculated using:
\begin{equation}
\begin{split}
\mathcal{I}(\hat{\mathcal{U}}_{k-1}) = \beta \text{Norm} (\mathcal{I}(\hat{\mathcal{U}}_{k-2})) + (1-\beta) \text{Norm} (\mathcal{I}(\mathcal{U}_{k-1})) \label{eq:norm}
\end{split}
\end{equation}
where $\beta \in [0, 1]$ and \text{Norm($\cdot$)} normalizes the importance scores to the [0, 1] range, ensuring consistency across models. The initial scores, $\mathcal{I}(\hat{\mathcal{U}}_1)$, are set equal to $\mathcal{I}(\mathcal{U}_1)$. Following this, the importance distribution for skill units up to task $\mathcal{T}_{k-1}$ is combined with that of the current task, $\mathcal{T}_{k}$, to facilitate the knowledge fusion process.

\subsection{Knowledge Fusion}
Following knowledge identification, the next critical phase is knowledge fusion, where knowledge within each skill unit is integrated into a cohesive framework. This process utilizes a sophisticated model averaging approach that accounts for various factors to optimize task performance.

Traditional coarse-grained model averaging operates under the assumption that all model weights hold equal importance for the training task \cite{kirkpatrick2017overcoming, eddine2023weighted}, typically implemented through the following iterative computation:
\begin{equation}
\hat{f}_k = \lambda \hat{f}_{k-1} + \left(1 - \lambda\right) f_k \label{eq:coarse}
\end{equation}

However, coarse-grained methods may overemphasize weights that are irrelevant to the current task, contaminating previously acquired task-specific knowledge and leading to forgetting. To overcome these limitations, we introduce two fine-grained averaging strategies that focus on skill units instead of the entire model. Our method distinguishes between task-shared and task-specific skill units, applying a weighted averaging technique to the parameters within each unit. This refined approach offers a tailored solution, addressing the unique requirements of each task more effectively.

\subsubsection{Static Weighted Fusion}
In the original {\ouralg} framework, this merging strategy initiates by establishing importance thresholds $\delta$ through quantiles, selecting the top 20\% of skill units based on importance scores. A skill unit $u_i^k$ is considered important (denoted as $(u_i^k)^+$) if its score $\mathcal{I}(u_i^k)$ exceeds $\delta_k$, and unimportant ($(u_i^k)^-$) otherwise.

The static weighted fusion strategy tailors parameter integration for each skill unit, according to its importance across different tasks. The method is defined as follows:
\begin{equation}
\hat{u}_i^k =
\begin{cases}
  \gamma \hat{u}_i^{k-1} + (1-\gamma) u_i^k, & \text{if $(\hat{u}_i^{k-1})^+$, $(u_i^k)^+$} \\

  \hat{u}_i^{k-1}, & \text{if $(\hat{u}_i^{k-1})^+$, $(u_i^k)^-$} \\
  
  u_i^k, & \text{if $(\hat{u}_i^{k-1})^-$, $(u_i^k)^+$} \\

  \frac{1}{2}(\hat{u}_i^{k-1} + u_i^k), & \text{if $(\hat{u}_i^{k-1})^-$, $(u_i^k)^-$} \\

\end{cases}
\label{eq:5}
\end{equation}
This approach performs the element-wise adjustment of parameters within each skill unit based on its relevance to previous and current tasks. The hyperparameter $\gamma$ is employed to control their influences.

\subsubsection{Adaptive Weighted Fusion}
As a sophisticated extension to the static method, which necessitated the manual setting of multiple hyperparameters, this adaptive strategy simplifies the process by automatically adjusting to various scenarios. This enhancement increases the efficiency and flexibility of the averaging process. Specifically, for a specific skill unit $u_i$, the weighting coefficients are determined by:
\begin{equation}
\begin{split}
\hat{\lambda}^{k-1}_{i} = \exp \left(\mathcal{I}(\hat{u}^{k-1}_{i}) / \tau\right), \lambda^{k}_{i} = \exp \left(\mathcal{I}(u^{k}_{i}) / \tau\right)
\label{eq:tau}
\end{split}
\end{equation}
where both $exp$ and temperature coefficient $\tau$ are scaled to the raw importance score. Then the updated model parameters are:
\begin{equation}
\begin{split}
\hat{u}_{i}^{k}=\left(\frac{\hat{\lambda}^{k-1}_{i}}{\hat{\lambda}^{k-1}_{i}+\lambda^{k}_{i}}\right) \cdot \hat{u}_{i}^{k-1}+ \left(\frac{\lambda^{k}_{i}}{\hat{\lambda}^{k-1}_{i}+\lambda^{k}_{i}}\right) \cdot u_{i}^{k} \label{eq:con}
\end{split}
\end{equation}

By applying Eq. (\ref{eq:5}) or Eq. (\ref{eq:con}) in our knowledge fusion phrase, it effectively addresses the challenges in CL:




\begin{itemize}[leftmargin=*,itemsep=2pt,topsep=0pt,parsep=0pt]
\item 
\textbf{Mitigating CF:} When $\mathcal{I}(\hat{u}^{k-1}_{i})$ is high and $\mathcal{I}(u^{k}_{i})$ is low, this indicates that the skill unit $u_i$ is more critical for previous tasks. 
According to Eq. (\ref{eq:5}) or (\ref{eq:con}), we maintain the previous task-specific knowledge in $\hat{u}^{k-1}_{i}$ untouched, preventing contamination from the less relevant $u^{k}_{i}$.

\item 
\textbf{Facilitating Forward KT:}
Conversely, if $\mathcal{I}(\hat{u}^{k-1}_{i})$ is low and $\mathcal{I}(u^{k}_{i})$ is high, it suggests that $u_i$ holds greater importance for the current task. 
As training initiates from the endpoint of the previous task, maintaining the integrity of parameters in $u^{k}_{i}$ leverages historically learned knowledge to enhance performance on the current task.


\item 
\textbf{Enabling Backward KT:}
When both $\mathcal{I}(\hat{u}^{k-1}_{i})$ and $\mathcal{I}(u^{k}_{i})$ are high, $u_{i}$ is vital for both previous and current tasks. Here, we integrate newly acquired knowledge into this task-shared skill unit to support backward KT.


\item 
\textbf{Handling Low Relevance:}
If both $\mathcal{I}(\hat{u}^{k-1}{i})$ and $\mathcal{I}(u^{k}{i})$ are low, indicating minimal relevance to any task, a simple averaging of parameters within this unit is adequate.


\end{itemize}

Knowledge fusion is conducted before starting a new task in CL, utilizing the averaged model for subsequent task initialization. Only the importance scores of $\hat{\mathcal{U}}_{k-1}$ and $\mathcal{U}_k$ are preserved for use between tasks, starting with $\hat{\mathcal{U}}_1 = \mathcal{U}_1$ estimated from $f_1$ on $D_1$. 
Detailed implementation of entire {\ouralg} algorithm is provided in Algorithm~\ref{alg:my_algorithm}.

\begin{algorithm}[t]
\caption{{\ouralg}}\label{alg:my_algorithm}
\begin{algorithmic}[1]
\renewcommand{\algorithmicrequire}{\textbf{Input:}}
\renewcommand{\algorithmicensure}{\textbf{Output:}}
\REQUIRE Dataset $\mathcal{D}_k$ for task $k$ = $1,\ldots,K$; initial pre-trained model $f_0$; hyperparameters $\alpha_1, \alpha_2, \beta, \tau$.

\STATE \textit{\# sequential tasks.}
\FOR {task $k$ = $1,\ldots,K$}  
    \STATE \textit{\# knowledge identification.} 
    \STATE Get $f_k$ and calculate $\mathcal{U}_k$ by Algorithm (\ref{alg:ipt});
    \IF{$k = 1$}
        \STATE \textit{\# initialization at beginning task.}
        \STATE $\hat{f}_1 \leftarrow f_1$, $\hat{\mathcal{U}}_1 \leftarrow \mathcal{U}_1$;
        
    \ELSE
        \STATE \textit{\# knowledge fusion.} 
        \FOR {skill unit $i$ = $1,\ldots,n$}  
        \STATE Calculate $\hat{u}_i^k$ by Eq. (\ref{eq:con});
        \ENDFOR
        \STATE Get the averaged model $\hat{f}_k$ based on $\hat{\mathcal{U}}_k$;
        \STATE Calculate accumulated importance score $\mathcal{I}(\hat{\mathcal{U}}_k)$ according to Eq. (\ref{eq:norm});
    \ENDIF   
\ENDFOR
\end{algorithmic} 
\end{algorithm}


\section{KIF with Memory Replay: KIF-M}
\begin{figure}[t]
  \centering
  \includegraphics[width=0.8\linewidth]{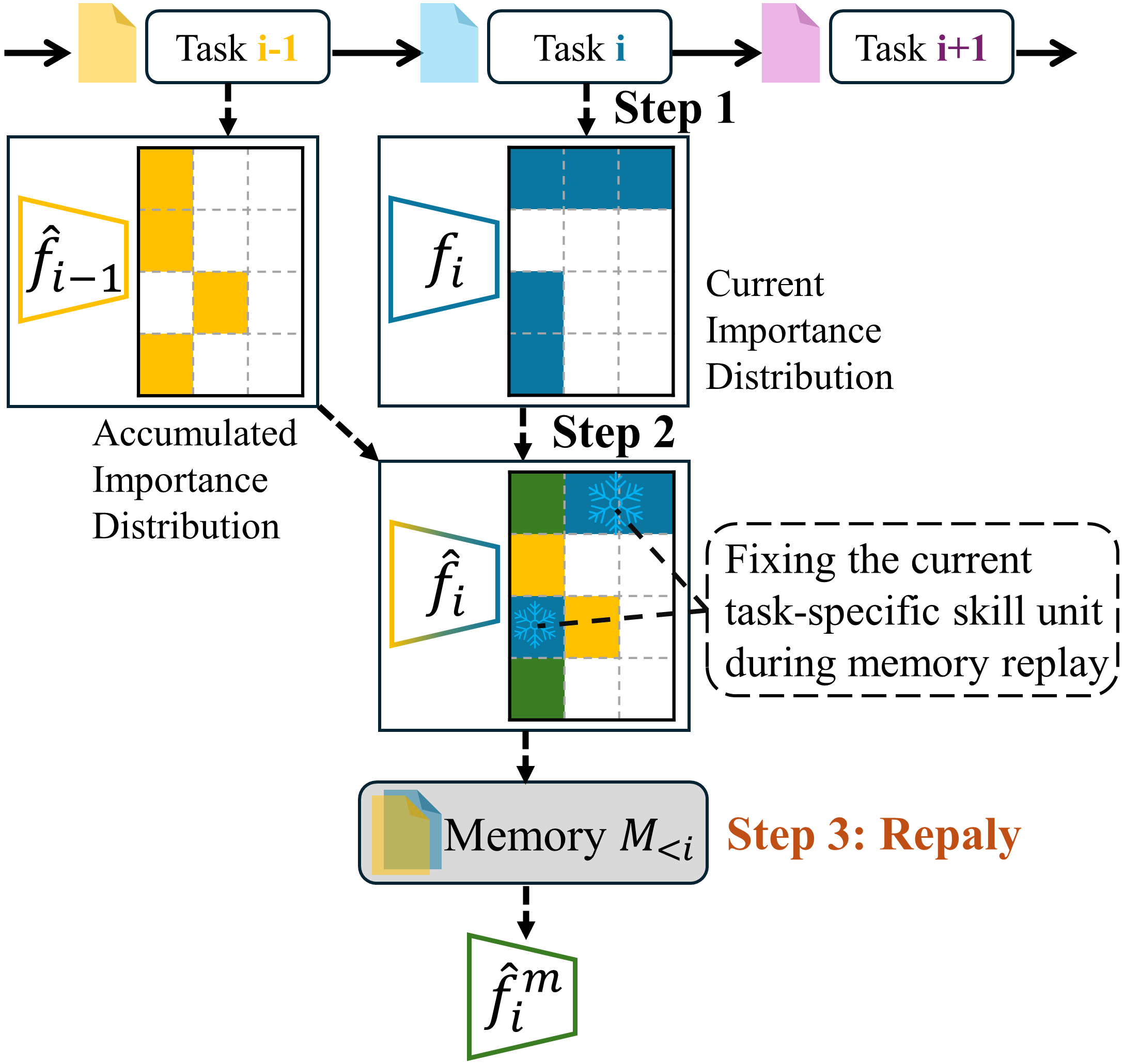}
  \caption{
  Overview of {\ouralgm}. By fixing the current task-specific skill unit during the replay phase, we can further enhance the performance of {\ouralg}.
  }
  \label{fig:methodM}
\end{figure}

To adapt {\ouralg} for scenarios where historical data are accessible, we introduce a memory replay-enhanced version, {\ouralgm}, designed to further improve model performance. The implementation process is shown in Figure \ref{fig:methodM}.

Following the training on task $\mathcal{T}_k$ using the knowledge identification and fusion, we obtain the model $\hat{f}_{k}$. Before the next task arrives, a replay stage is introduced, where the model $\hat{f}_{k}$ is fine-tuned using historical data stored in the memory buffer $\mathcal{M}_{<k}$. This process results in a new model $\hat{f}^{m}_{k}$, which helps recover forgotten knowledge.

Specifically, during the replay stage, we avoid fine-tuning the parameters of all skill units with historical data, as this could potentially degrade the model's performance on the current task. Instead, we maintain the integrity of the current task-specific skill units and selectively fine-tune the remaining parameters. This approach achieves a better balance between retaining previous knowledge and excelling in new tasks.

\section{Experiments and Analysis}\label{sec:exp}
\subsection{Experimental Setup}
\subsubsection{Datasets}
We utilize the SuperNI Benchmark~\cite{wang2022super}, a comprehensive benchmark designed to evaluate diverse NLP tasks using expert-written instructions. This benchmark facilitates thorough evaluation in more practical settings for the CL of LLMs. It includes tasks in dialogue generation, information extraction, question answering, summarization, and sentiment analysis.
Following the established CL setup~\cite{zhao2024sapt}, three tasks are chosen from each type, creating a sequence of 15 tasks in total for evaluation. For each task, 1,000 instances from the dataset are randomly selected for training, and 100 instances are used for validation and testing.

Additionally, we utilize the Long Sequence Benchmark~\cite{razdaibiedina2022progressive}, which consists of 15 classification datasets tailored for CL. Consistent with previous work~\cite{wang2023orthogonal}, we randomly select 1,000 samples per task for training and reserve 500 samples per class for validation and testing.
The experiments include two different task orders for each benchmark. 


\newcommand{\tabincell}[2]{\begin{tabular}{@{}#1@{}}#2\end{tabular}}
\begin{table*}[t]
\caption{The overall results on two continual learning benchmarks with T5-large model. All results are averaged over two different orders of each benchmark. 
}
\centering
\scalebox{1.2}{
\begin{tabular}{l|c|ccc|ccc}
\toprule
\multirow{2}*{\tabincell{c}{Method}}&   \multirow{2}*{\tabincell{c}{Memory-Replay}}  & \multicolumn{3}{c|}{SuperNI Benchmark} & \multicolumn{3}{c}{Long Sequence Benchmark} \\
&    & AP$\uparrow$ &  FWT$\uparrow$&  BWT$\uparrow$ & AP$\uparrow$&  FWT$\uparrow$&  BWT$\uparrow$ \\
\midrule
\rule{0pt}{4pt}SeqLoRA  &  \multirow{9}*{\tabincell{c}{\xmark}}   & 6.43 & 1.58 & -30.94 & 9.72 & 6.81& -73.37 \\
\rule{0pt}{8pt}IncLoRA  &   & 19.12   &  2.03 &  - 31.24 & 62.50 & 2.62  & -15.34  \\
\rule{0pt}{8pt}ProgPrompt~\cite{razdaibiedina2022progressive} &    & 3.34 & 5.29 & -33.18 & 7.98 & 6.63 & -66.71\\
\rule{0pt}{8pt}EWC~\cite{kirkpatrick2017overcoming}  &   & 17.46  &  4.20 &  -28.61  & 45.45 & 3.73  & -25.93   \\
\rule{0pt}{8pt}L2P~\cite{wang2022learning}  & & 12.73 & 1.14 & -7.95  & 57.98 & 8.36 & -16.63 \\
\rule{0pt}{8pt}LFPT5~\cite{qin2021lfpt5}   &    & 24.76 & 7.46 & -24.41 &  67.01 & 9.48 & -12.80 \\

\rule{0pt}{8pt}O-LoRA~\cite{wang2023orthogonal} &    & 25.89 & 8.14 &-24.59 &  69.24 & 10.15& -4.05\\
\rowcolor[gray]{0.9}
\rule{0pt}{8pt}\textbf{{\ouralg} (ours)} &    & 26.41 &\textbf{11.78}  &-18.55 &  70.71 & 11.80& -3.67\\
\rowcolor[gray]{0.9}
\rule{0pt}{8pt}\textbf{{\ouralgn} (ours)} &   &  \textbf{27.43}  & 11.02& \textbf{-16.91} & \textbf{72.29} & \textbf{12.89} & \textbf{-3.04} \\

\midrule
\rule{0pt}{8pt}Replay  &  \multirow{4}*{\tabincell{c}{\cmark}}  & 35.37 & 2.35& -15.79& 70.98 & 3.28& -15.42 \\
\rule{0pt}{8pt}SAPT~\cite{zhao2024sapt} &    & 51.54 & 8.88 &\textbf{-0.57}& 82.02 & 9.86 &-1.25\\
\rowcolor[gray]{0.9}
\rule{0pt}{8pt}\textbf{{\ouralgm} (ours)} & &  52.12 & 11.13& -1.31 & 84.21 & \textbf{12.32} & -1.08  \\
\rowcolor[gray]{0.9}
\rule{0pt}{8pt}\textbf{KIFLoRA-M (ours)} & &  \textbf{53.01}  & \textbf{11.21}& -0.81 & \textbf{84.36} & 11.71 & \textbf{-0.98}  \\


\bottomrule
\end{tabular}}
\label{tbl:result}
\end{table*}

\subsubsection{Evaluation Metrics}
We denote $a_{j,i}$ as the testing performance (Accuracy for classification task and Rouge-L \cite{lin2004rouge} for others) on the $i$-th test set of task right after training on $j$-th task.
The performance of CL is assessed using three metrics:
\begin{equation}
   \mathbf{AP} =\frac{1}{K} \sum_{i=1}^{K} a_{K, i}
\end{equation}
\begin{equation}
    \mathbf{FWT} = \frac{1}{K-1} \sum\limits_{i=2}^{K} a_{i-1, i}
\end{equation}
\begin{equation}
\mathbf{BWT} = \frac{1}{K-1} \sum\limits_{i=1}^{K-1} a_{K, i}-a_{i, i}
\end{equation}

\textbf{Average Performance (AP)} \cite{de2021continual} calculates the average performance across all tasks after training on the final task.
\textbf{Forward Transfer (FWT)} \cite{lopez2017gradient} evaluates a model's generalization ability by measuring the averaged zero-shot performance.
\textbf{Backward Transfer (BWT)} \cite{ke2022continual} assesses the impact of new learning on previous tasks. Negative BWT indicates the model lost some previously acquired knowledge.

\subsubsection{Baselines}
We evaluate {\ouralg} against the following PEFT-based CL baseline methods:
\textbf{\emph{SeqLoRA:}} sequentially trains the LoRA on the task orders.
\textbf{\emph{IncLoRA:}} incremental learning of new LoRA parameters on a sequential series of tasks.
\textbf{\emph{Replay:}} replays real samples from old tasks when learning new tasks to avoid forgetting.
\textbf{\emph{EWC \cite{kirkpatrick2017overcoming}:}} finetune the model with a regularization loss designed to minimize updates to parameters critical to previously learned tasks.
\textbf{\emph{L2P \cite{wang2022learning}:}} uses the input to dynamically select and update prompts from a fixed prompt pool.
\textbf{\emph{LFPT5 \cite{qin2021lfpt5}:}} continuously trains a soft prompt for each task with generative replay.
\textbf{\emph{Prog-Prompt \cite{razdaibiedina2022progressive}:}} sequentially concatenates previous learned prompts to the current one during the training and testing time.
\textbf{\emph{O-LoRA \cite{wang2023orthogonal}:}} learns tasks in distinct LoRA subspaces, ensuring orthogonality, and aggregates all LoRA weights for testing.
\textbf{\emph{SAPT \cite{zhao2024sapt}:}} leverages pseudo samples and a shared attention framework to align PEFT block learning and selection.

Table \ref{tbl:compare} compares our {\ouralg} with other baselines, highlighting three key advantages: data privacy-friendliness, model parameter efficiency, and improved generalization capabilities.
To clarify the specific identification and fusion techniques used in {\ouralg} and its variants:
\textbf{\emph{{\ouralg}:}} our foundational framework incorporates matrix-level skill units, utilizes a first-order gradient-based metric for knowledge identification (Eq. \ref{eq:1}), and employs static weighted fusion (Eq. \ref{eq:5}).
\textbf{\emph{{\ouralgn}:}} an extension of {\ouralg} that introduces LoRA-tailored skill units, adopts a second-order gradient approximation metric (Eq. \ref{eq:ipt}), and features adaptive weighted fusion (Eq. \ref{eq:con}).
\textbf{\emph{{\ouralgm} and KIFLoRA-M:}} these enhancements of {\ouralg} and {\ouralgn} integrate memory replay to further boost performance.

\begin{table}
\caption{The comparison between {\ouralg} and other CL methods.
Specifically, \textbf{RF} indicates whether the method is rehearsal-free. \textbf{PE} indicates whether the method is parameter efficient. \textbf{UT} indicates whether the method can be applied to solve unseen tasks. \textbf{KT} indicates whether the method enables knowledge transfer.}
\centering
\scalebox{1}{
\begin{tabular}{lcccc}
\toprule
Method & RF & PE & UT & KT\\
\midrule
\rule{0pt}{6pt} EWC \cite{kirkpatrick2017overcoming}& \checkmark  &   &  & \checkmark   \\
\rule{0pt}{8pt} OGD \cite{farajtabar2020orthogonal} &  \checkmark&   & \checkmark     \\
\rule{0pt}{8pt} LFPT5 \cite{qin2021lfpt5} &  & \checkmark  & & \checkmark  \\
\rule{0pt}{8pt} L2P \cite{wang2022learning}&  \checkmark &  \checkmark  &    \\
\rule{0pt}{8pt} EIP \cite{wang2023rehearsal} &  \checkmark &  \checkmark &    \\
\rule{0pt}{8pt} O-LoRA \cite{wang2023orthogonal}& \checkmark   &\checkmark   & \checkmark  &    \\
\rule{0pt}{8pt} MoCL \cite{wang2024rehearsal}& \checkmark   &  & \checkmark  &  \checkmark  \\
\rule{0pt}{8pt} SAPT \cite{zhao2024sapt} &    &\checkmark   & \checkmark  & \checkmark    \\

\midrule
\rowcolor[gray]{0.9}
\rule{0pt}{8pt} \textbf{{\ouralg}}  & \checkmark   &\checkmark   & \checkmark  & \checkmark  \\

\bottomrule
\end{tabular}}
\label{tbl:compare}
\end{table}

\subsubsection{Implementation Details}
We utilize two distinct language model architectures:
the encoder-decoder T5 model \cite{raffel2020exploring} and the decoder-only LLaMA model \cite{touvron2023llama}.
In {\ouralg}, the hyperparameters $\alpha_1$ and $\alpha_2$ in Eq. (\ref{eq:I}) and Eq. (\ref{eq:U}) are set to 0.85. We set $\beta$ in Eq. (\ref{eq:norm}) to 0.7 and $\tau$ in Eq. (\ref{eq:tau}) to 0.15.
Following \cite{sun2019lamol}, we allocate 2\% of the original training set volume for replay samples in all replay-based baseline methods.
Hyperparameters are as follows: 
\begin{itemize}[leftmargin=*,itemsep=2pt,topsep=0pt,parsep=0pt]
\item T5-base (220M), T5-large (780M), and T5-XL (3B): Training was conducted with a learning rate of 3e-4, batch size of 8, maximum input length of 512, maximum target length of 128, a rank of 8, targeting modules [q, v] and 10 epochs.

\item LLaMA (7B): With a learning rate of 3e-4, a batch size of 128, a cutoff length of 512, and 10 epochs. LoRA settings were rank = 8, alpha = 16, dropout = 0.05, targeting modules [q\_proj, v\_proj]. For testing, settings included temperature = 0.02, top\_p = 0, top\_k = 1, max new tokens = 128.

\end{itemize}


\subsection{Main Results}
The overall CL results of various methods using the same T5-large backbone are summarized in Table \ref{tbl:result}.

Firstly, {\ouralg} and its variants consistently demonstrate superior CL performance by effectively facilitating knowledge transfer. Compared to previous replay-free methods like IncLoRA and O-LoRA, our methods excel in addressing the critical challenges of CF and KT, achieving the highest AP, FWT, and BWT when learning tasks sequentially.

Furthermore, our enhanced version, {\ouralgn}, significantly outperforms {\ouralg}. Specifically, {\ouralgm} achieves average gains of 1.3\% in AP and 0.3\% in BWT, underscoring the effectiveness of enhancements to each framework component. Notably, even without memory replay, {\ouralgm} surpasses the performance of vanilla repla on the Long Sequence Benchmark, particularly in BWT.

For the memory-based methods, to ensure a fair comparison, we adopted the same data synthesis-based generative replay approach used in SAPT, ensuring that the amount of data used in memory replay is consistent across methods.
The results show that both {\ouralgm} and KIFLoRA-M surpass the best SAPT models, further proving the versatility and robustness of our framework. For subsequent experimental analyses and comparisons, we will focus on {\ouralgn}, the most effective memory-free version, for further evaluations.

\begin{figure*}[t]
  \centering
  \subfigure[Results on the SuperNI benchmark]{\includegraphics[width=0.49\linewidth]{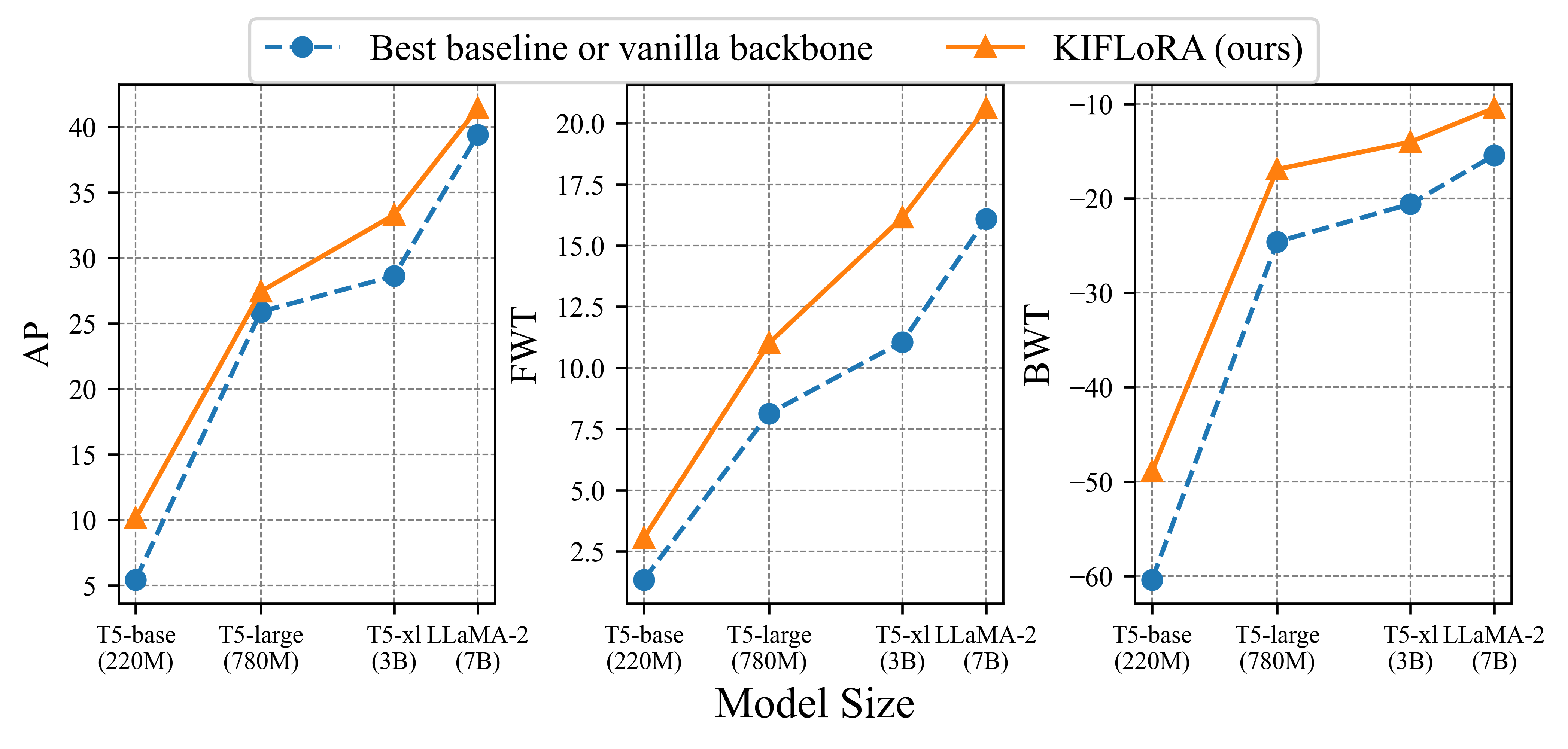}}
  \subfigure[Results on the Long Sequence benchmark]{\includegraphics[width=0.49\linewidth]{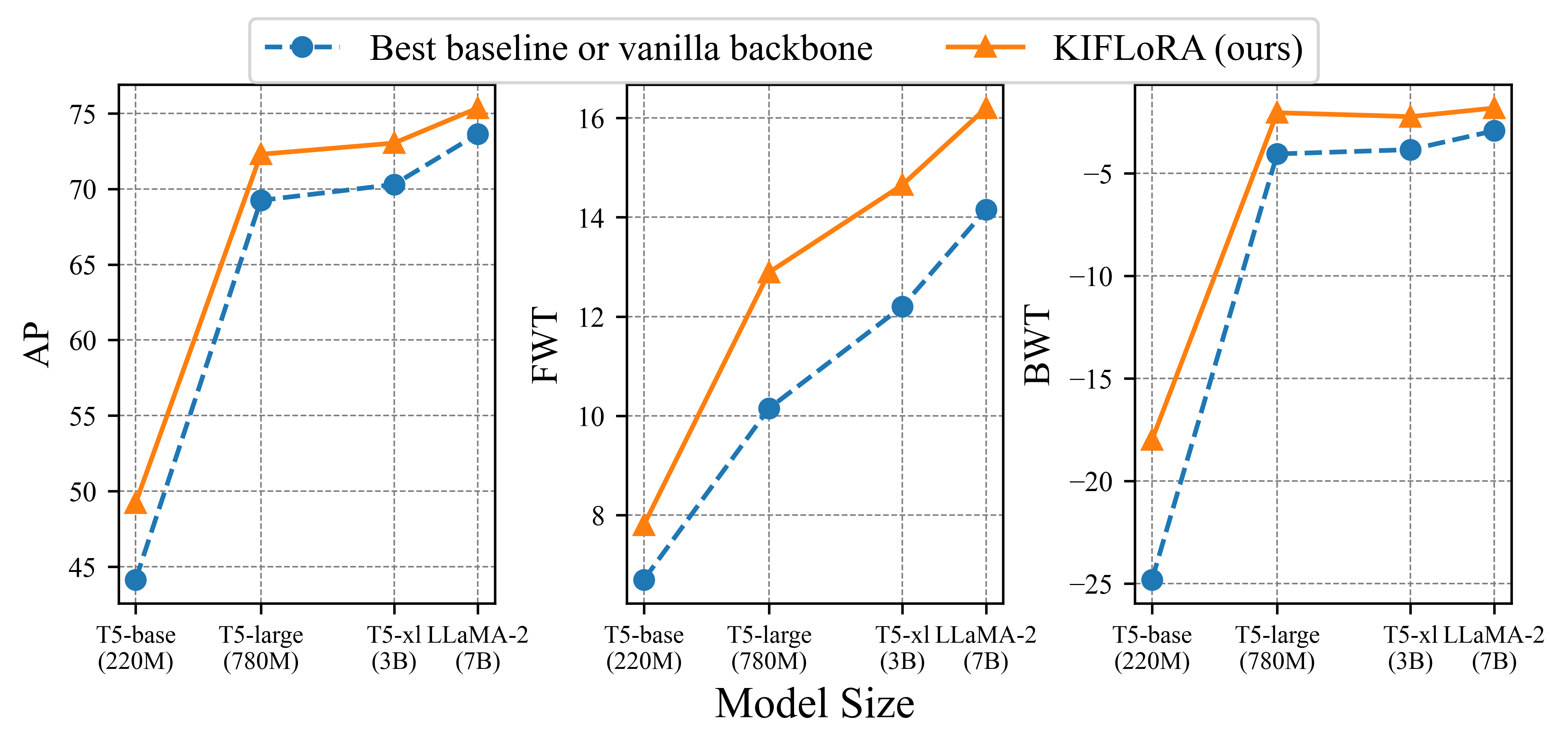}}
  \caption{Performance of {\ouralgn} across various backbones.
  }
  \label{fig:different_size}
\end{figure*}

Secondly, {\ouralgn} consistently demonstrates superior performance across various backbones. To further validate our framework's effectiveness, we conducted experiments using a range of parameter-level backbones. As depicted in Figure \ref{fig:different_size}, performance improvements are evident with increasing model size. 
For instance, in T5-XL, {\ouralgn} boosts the AP metric from 28.6\% to 33.3\%, and shows substantial enhancements in FWT and BWT, rising from 11.1\% to 16.1\% and improving from -20.6\% to -14.0\%, respectively. These results further validate the robust generalization ability of our method.


\begin{figure}[t]
  \centering
  \includegraphics[width=0.8\linewidth]{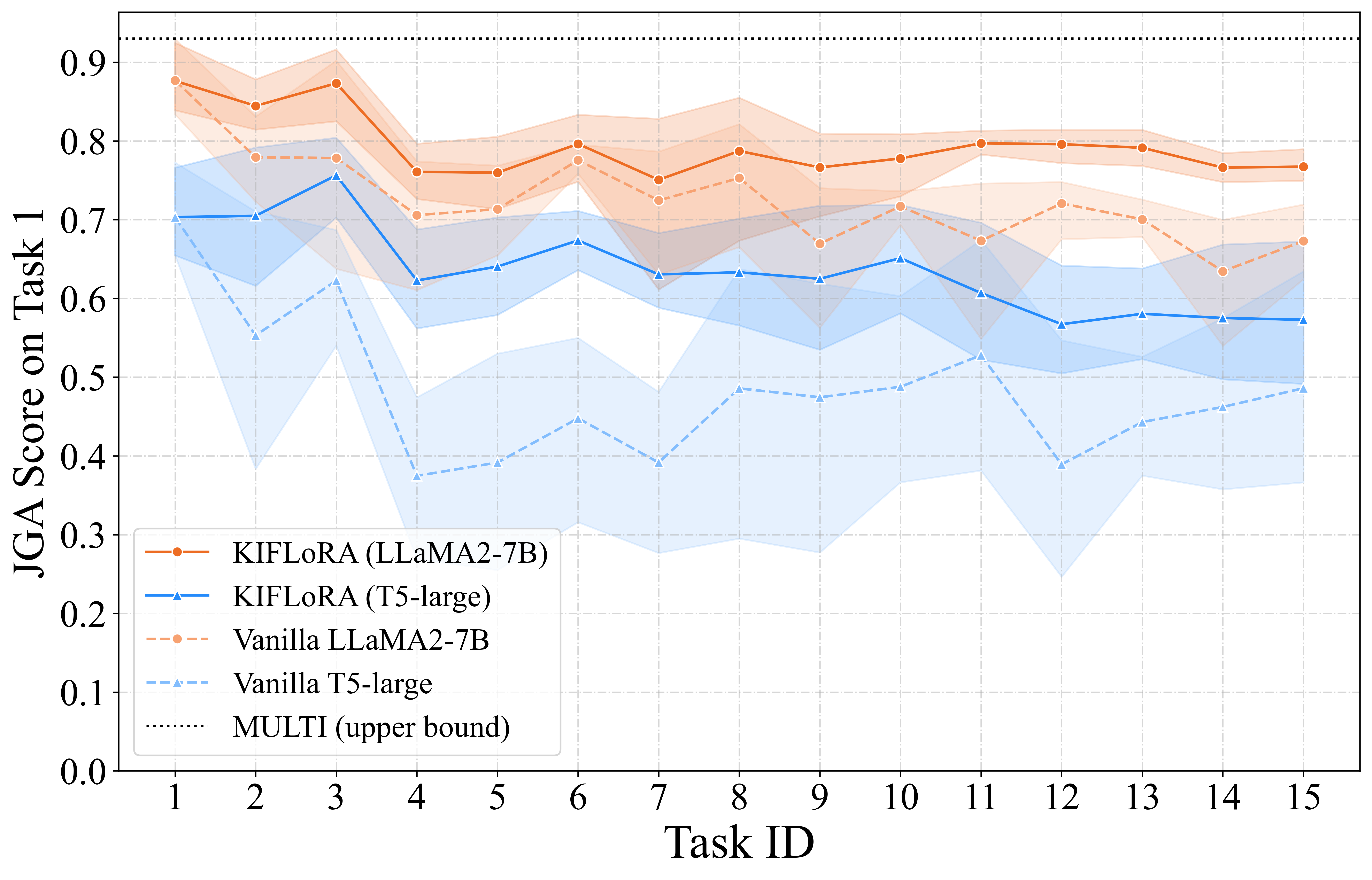}
  \caption{Performance trajectory of Task 1 on Long Sequence Benchmark throughout the continual learning process.
  }
  \label{fig:forgetting}
\end{figure}

Thirdly, our knowledge fusion technique effectively counters catastrophic forgetting. To rigorously assess our model's ability to mitigate forgetting, we evaluated its performance on the initial task after training on subsequent tasks using the Long Sequence Benchmark. Figure \ref{fig:forgetting} illustrates that {\ouralgn} significantly reduces the rate of forgetting, with an average performance decline of only 10\% after training on the final task. In contrast, vanilla backbones suffer a substantial average performance drop of 22\%, highlighting our method's enhanced capacity to preserve learned knowledge.

\subsection{Performance on Unseen Tasks}
Building on prior research \cite{zhao2024sapt}, we selected three tasks from each category to evaluate our method's cross-task generalization capabilities, a critical measure for CL algorithms. The results, indicated by the average Rouge-L scores in Table \ref{unseen}, highlight the effectiveness of our approach. T5-ZS refers to the zero-shot task adaptation methods employed. {\ouralgn} exhibited the best performance among all methods tested, a success largely due to its proficient identification and management of task-specific and shared parameters.

\begin{table}[]
\caption{Results on unseen tasks using the T5-Large backbone.}
\centering
\scalebox{1}{
\begin{tabular}{
l |
c 
r 
r 
r
r|
r}
\toprule
& \multicolumn{5}{c|}{\textbf{Unseen Tasks}} & \multirow{2}{*}{\textbf{Avg.}} \\
\textbf{}        & \textbf{Dialog}     & \multicolumn{1}{c}{\textbf{IE}}      & \multicolumn{1}{c}{\textbf{QA}}      & \textbf{Sum} & \multicolumn{1}{c|}{\textbf{SA}}  &   \\ \midrule
T5-ZS  &    7.49         &     6.70      &   4.28      &  12.14  &   4.54  & 
 7.03 \\
O-LoRA   &    4.39     &  9.89    &  25.38  &  8.26  &  \textbf{50.41}  & 19.67  \\
LFPT5   &    6.96     &  \textbf{35.32}    &  35.00  &  13.26  &  21.51  & 22.41  \\
\midrule
\rowcolor[gray]{0.9}
{\ouralgn}   &  \textbf{10.32}     &  31.34     & \textbf{37.13}      & \textbf{14.20} & 47.17 & \textbf{28.03}         \\
\bottomrule
\end{tabular}
}
\label{unseen}
\end{table}

\subsection{Visualization of Skill Units}
\begin{figure*}[t]
  \centering
  \includegraphics[width=0.9\linewidth]{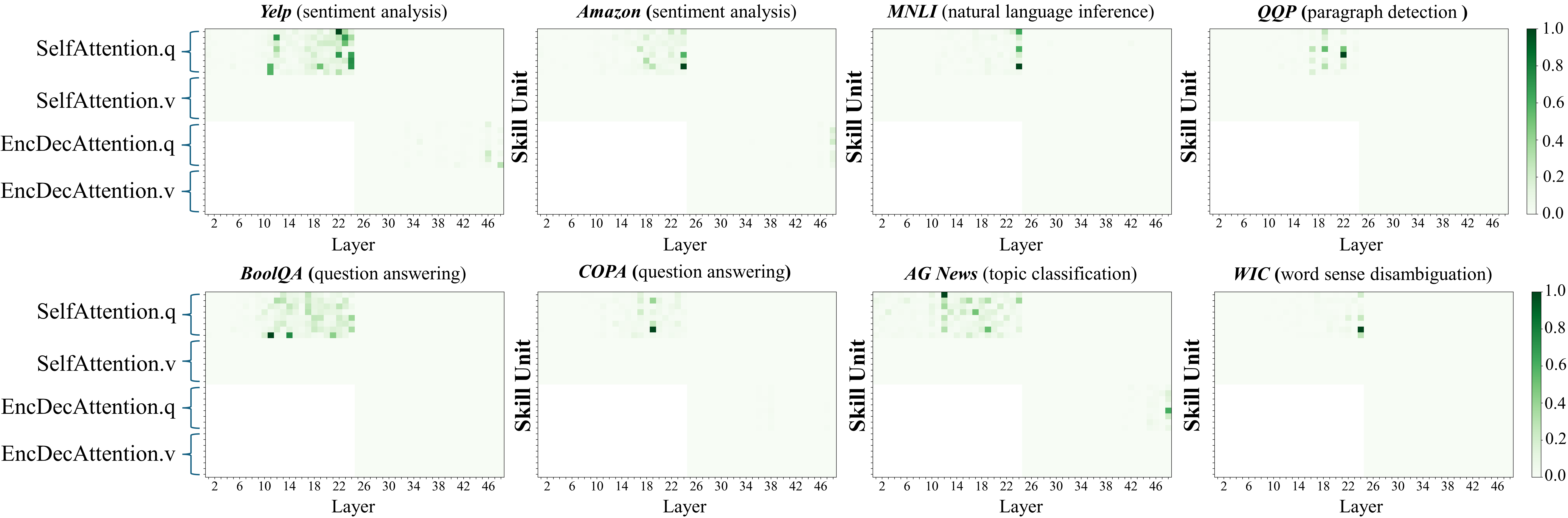}
  \caption{Visualization of importance scores for LoRA-tailored skill units across different tasks on T5-large within the Long Sequence Benchmark.
  }
  \label{fig:heat_map_long}
\end{figure*}

\begin{figure*}[t]
  \centering
  \includegraphics[width=0.9\linewidth]{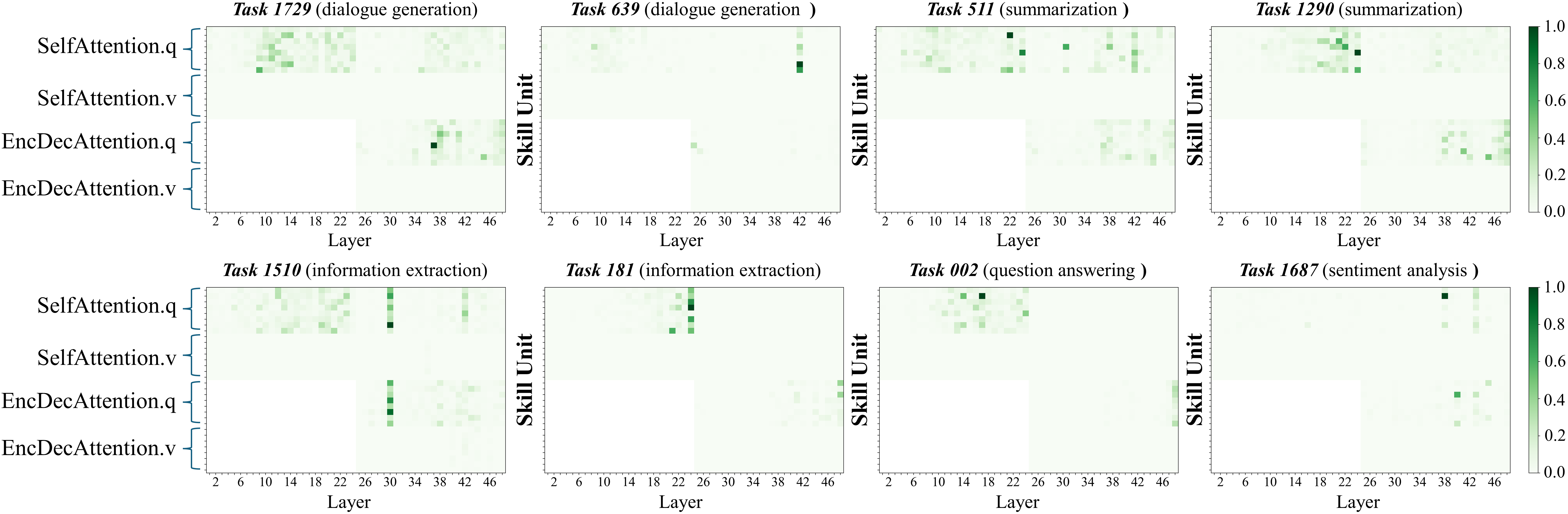}
  \caption{Visualization of importance scores for LoRA-tailored skill units across different tasks on T5-large within the SuperNI Benchmark.
  }
  \label{fig:heat_map_superni}
\end{figure*}

We visualized the distribution of importance scores for skill units across tasks on the T5-large model, as shown in Figure \ref{fig:heat_map_long} and Figure \ref{fig:heat_map_superni}, leading to several critical insights:

\begin{itemize}[leftmargin=*,itemsep=2pt,topsep=0pt,parsep=0pt]
\item 
There is noticeable variability in the importance of skill units for the same task, with important skill units comprising only a small fraction of all trainable LoRA parameters. 

\item 

The distribution of important skill units is task-dependent, illustrating the presence of both task-shared and task-specific parameters. This observation substantiates the feasibility of the design motivations behind the {\ouralg} framework.

\item 

For classification tasks, such as those in the Long Sequence Benchmark (Figure \ref{fig:heat_map_long}), the skill units in the encoder, particularly the lower layers closer to the model input, play a more pivotal role. Conversely, for generative tasks, like dialogue generation and summarization in the SuperNI benchmark (Figure \ref{fig:heat_map_superni}), both encoder and decoder units are crucial, impacting both lower and upper layers of the network.

\item 

Within each layer, the importance of the Query (Q) matrices in the attention mechanism consistently exceeds that of the Value (V) matrices. 

\end{itemize}

\begin{table}
\caption{Ablation Study for Importance-Aware Knowledge Identification.}
\centering
\scalebox{1}{
\begin{tabular}{lccc}
\toprule
Method & AP & FWT & BWT\\
\midrule
\rule{0pt}{6pt} vanilla T5-large &54.10 & 3.32 & -29.63 \\
\midrule
\rule{0pt}{8pt} $s\left( \cdot \right) = I\left( \cdot \right)$ &  70.48 & 10.39  & -3.81 \\
\rule{0pt}{8pt} $s\left( \cdot \right) = \left|\nabla_{w_{i j}} \mathcal{L}\right|$ & 68.82  &  10.80 & -6.22 \\

\rowcolor[gray]{0.9}
\rule{0pt}{8pt} {\ouralgn} (ours) & \textbf{72.29}  & \textbf{12.89}  & \textbf{-2.04} \\

\bottomrule
\end{tabular}}
\label{tbl:ablation_ipt}
\end{table}

\subsection{Ablation Study}
In this section, we assess the impact of importance-aware knowledge identification, fine-grained knowledge fusion, and memory size on model performance. Detailed analysis of hyperparameter sensitivity is provided in the Appendix.

\subsubsection{Effect of the Proposed Importance Metric in Knowledge Identification} 
We explore alternative importance scoring methods to validate the effectiveness of our approach: (i) To evaluate the impact of using moving averages on trajectory gradients, we modify $s(\cdot)$ in Eq. (\ref{eq:3}) to focus solely on sensitivity, as outlined in Eq. (\ref{eq:ipt}); (ii) To verify the validity of our proposed approximate second-order gradient importance metric in Eq. (\ref{eq:3}), we implement a first-order Taylor expansion as described in Eq. (\ref{eq:1}). The comparative results, presented in Table \ref{tbl:ablation_ipt}, show that using moving averages for importance scoring substantially enhances model performance. Additionally, compared to the first-order gradient approximation, our novel second-order gradient method results in performance improvements of up to 3.4\%, 2.1\%, and 4.2\% across the evaluated metrics. These findings underscore the significant role of accurate knowledge identification.

\subsubsection{Effect of Adaptive Knowledge Fusion}
We evaluate our fine-grained, skill unit-level adaptive weighted fusion against two coarse-grained strategies: (i) Weight-Ensemble, which uniformly averages the LoRA parameters using a global weight as per Eq. (\ref{eq:coarse}), and (ii) Exponential Moving Average (EMA) \cite{szegedy2016rethinking}, which computes a running average of parameters at each fine-tuning iteration.

According to Table \ref{tbl:ablation_avg}, Weight-Ensemble significantly enhances the performance of the vanilla model, demonstrating the advantages of coarse-grained averaging in CL. While EMA generally outperforms Weight-Ensemble, it does not match the effectiveness of our fine-grained approach. EMA's frequent parameter adjustments within the same task can lead to suboptimal outcomes. In contrast, our method, which strategically averages weights only after the completion of each task, boosts computational efficiency and model performance.

\begin{table}[t]
\caption{Ablation Study for Fine-Grained Knowledge Fusion.}
\centering
\scalebox{1}{
\begin{tabular}{lccc}
\toprule
Method & AP & FWT & BWT\\
\midrule
\rule{0pt}{6pt} vanilla T5-large &54.10 & 3.32& -29.63 \\
\midrule
\rule{0pt}{8pt} Weight-Ens.& 63.28  & 7.71  & -11.82 \\
\rule{0pt}{8pt} EMA & 62.76  & 8.23  &  -9.80\\
\rowcolor[gray]{0.9}
\rule{0pt}{8pt} {\ouralgn} (ours) & \textbf{72.29}  & \textbf{12.89}  & \textbf{-3.04} \\
\bottomrule
\end{tabular}}
\label{tbl:ablation_avg}
\end{table}

\subsubsection{The Effect of Memory Size}
We examine how varying the memory size affects the performance of Replay, {\ouralgm}, and KIFLoRA-M. By adjusting the memory size per task \(|M|\) to {2\%, 5\%, 10\%, 50\%}, we summarized the results in Table \ref{tbl:ablation_memory}. As expected, increasing the memory size generally enhances the performance of all methods, with Replay showing the most significant improvement. This is predictable as Replay heavily depends on memory to mitigate catastrophic forgetting. When the memory size is substantially increased, Replay approaches the functionality of multi-task learning, which, while effective, incurs considerable storage and computational costs.

Conversely, {\ouralgm} and KIFLoRA-M utilize knowledge fusion strategies to effectively maintain parameters that store historical knowledge. Consequently, while increasing memory size does boost their performance, the improvements are less dramatic as the memory size grows.

\begin{table}[t]
\caption{Ablation Study on Memory Size.
}
\centering
\scalebox{1}{
\begin{tabular}{lcccc}
\toprule
\multirow{2}*{\tabincell{c}{ }} & \multicolumn{4}{c}{Memory Size}\\
\cmidrule(lr){2-5}
 & 2\% & 5\% & 10\% & 50\%\\
\midrule
\rule{0pt}{6pt} Replay & 71.2 & 72.4 & 73.8 & 76.1   \\
\rowcolor[gray]{0.9}
\rule{0pt}{8pt} {\ouralgm} & 74.2 & 75.1  & 75.7 & 79.2 \\
\rowcolor[gray]{0.9}
\rule{0pt}{8pt} KIFLoRA-M & \textbf{74.3}  & \textbf{75.4}  & \textbf{76.2} & \textbf{79.8}\\
\bottomrule
\end{tabular}}
\label{tbl:ablation_memory}
\end{table}

\section{Conclusion}
In this paper, we introduced the novel Knowledge Identification and Fusion ({\ouralg}) framework for language model continual learning. {\ouralgn} employs importance-aware knowledge identification and fine-grained knowledge fusion to effectively differentiate between task-specific and shared knowledge within each skill unit, mitigating forgetting while enhancing knowledge transfer. The framework is highly generalizable and extensible, allowing integration with memory replay methods to boost performance. Extensive testing shows that {\ouralgn} excels in preserving past knowledge and performing well in new tasks, surpassing previous state-of-the-art approaches.
Comprehensive experiments demonstrate demonstrate our framework's exceptional ability.

Despite significant improvements in efficiency and performance in large language model continual learning, certain limitations persist. For instance, the choice of importance thresholds following the determination of importance distribution can impact the effectiveness of knowledge fusion. Dynamically selecting these thresholds based on data distribution may yield a more precise classification of task-shared and task-specific parameters, further enhancing performance. Additionally, merging the knowledge identification and fusion phases could allow for ongoing parameter consolidation based on their importance during model training. This integrated approach would facilitate more flexible adaptations and better address scenarios in online continual learning.




%

\bibliographystyle{IEEEtran}
\bibliography{references}

\begin{thebibliography}{10}
\providecommand{\url}[1]{#1}
\csname url@samestyle\endcsname
\providecommand{\newblock}{\relax}
\providecommand{\bibinfo}[2]{#2}
\providecommand{\BIBentrySTDinterwordspacing}{\spaceskip=0pt\relax}
\providecommand{\BIBentryALTinterwordstretchfactor}{4}
\providecommand{\BIBentryALTinterwordspacing}{\spaceskip=\fontdimen2\font plus
\BIBentryALTinterwordstretchfactor\fontdimen3\font minus \fontdimen4\font\relax}
\providecommand{\BIBforeignlanguage}[2]{{%
\expandafter\ifx\csname l@#1\endcsname\relax
\typeout{** WARNING: IEEEtran.bst: No hyphenation pattern has been}%
\typeout{** loaded for the language `#1'. Using the pattern for}%
\typeout{** the default language instead.}%
\else
\language=\csname l@#1\endcsname
\fi
#2}}
\providecommand{\BIBdecl}{\relax}
\BIBdecl

\bibitem{brown2020language}
T.~Brown, B.~Mann, N.~Ryder, M.~Subbiah, J.~D. Kaplan, P.~Dhariwal, A.~Neelakantan, P.~Shyam, G.~Sastry, A.~Askell \emph{et~al.}, ``Language models are few-shot learners,'' \emph{Advances in neural information processing systems}, vol.~33, pp. 1877--1901, 2020.

\bibitem{de2021continual}
M.~De~Lange, R.~Aljundi, M.~Masana, S.~Parisot, X.~Jia, A.~Leonardis, G.~Slabaugh, and T.~Tuytelaars, ``A continual learning survey: Defying forgetting in classification tasks,'' \emph{IEEE transactions on pattern analysis and machine intelligence}, vol.~44, no.~7, pp. 3366--3385, 2021.

\bibitem{wang2024comprehensive}
L.~Wang, X.~Zhang, H.~Su, and J.~Zhu, ``A comprehensive survey of continual learning: Theory, method and application,'' \emph{IEEE Transactions on Pattern Analysis and Machine Intelligence}, 2024.

\bibitem{long2018transferable}
M.~Long, Y.~Cao, Z.~Cao, J.~Wang, and M.~I. Jordan, ``Transferable representation learning with deep adaptation networks,'' \emph{IEEE transactions on pattern analysis and machine intelligence}, vol.~41, no.~12, pp. 3071--3085, 2018.

\bibitem{mccloskey1989catastrophic}
M.~McCloskey and N.~J. Cohen, ``Catastrophic interference in connectionist networks: The sequential learning problem,'' in \emph{Psychology of learning and motivation}.\hskip 1em plus 0.5em minus 0.4em\relax Elsevier, 1989, vol.~24, pp. 109--165.

\bibitem{ke2021achieving}
Z.~Ke, B.~Liu, N.~Ma, H.~Xu, and L.~Shu, ``Achieving forgetting prevention and knowledge transfer in continual learning,'' \emph{Advances in Neural Information Processing Systems}, vol.~34, 2021.

\bibitem{hu2021lora}
E.~J. Hu, Y.~Shen, P.~Wallis, Z.~Allen-Zhu, Y.~Li, S.~Wang, L.~Wang, and W.~Chen, ``Lora: Low-rank adaptation of large language models,'' \emph{arXiv preprint arXiv:2106.09685}, 2021.

\bibitem{ding2022delta}
N.~Ding, Y.~Qin, G.~Yang, F.~Wei, Z.~Yang, Y.~Su, S.~Hu, Y.~Chen, C.-M. Chan, W.~Chen \emph{et~al.}, ``Delta tuning: A comprehensive study of parameter efficient methods for pre-trained language models,'' \emph{arXiv preprint arXiv:2203.06904}, 2022.

\bibitem{shu2023omni}
Y.~Shu, Z.~Cao, J.~Gao, J.~Wang, S.~Y. Philip, and M.~Long, ``Omni-training: bridging pre-training and meta-training for few-shot learning,'' \emph{IEEE Transactions on Pattern Analysis and Machine Intelligence}, 2023.

\bibitem{rusu2016progressive}
A.~A. Rusu, N.~C. Rabinowitz, G.~Desjardins, H.~Soyer, J.~Kirkpatrick, K.~Kavukcuoglu, R.~Pascanu, and R.~Hadsell, ``Progressive neural networks,'' \emph{arXiv preprint arXiv:1606.04671}, 2016.

\bibitem{wang2023rehearsal}
Z.~Wang, Y.~Liu, T.~Ji, X.~Wang, Y.~Wu, C.~Jiang, Y.~Chao, Z.~Han, L.~Wang, X.~Shao \emph{et~al.}, ``Rehearsal-free continual language learning via efficient parameter isolation,'' in \emph{Proceedings of the 61st Annual Meeting of the Association for Computational Linguistics (Volume 1: Long Papers)}, 2023, pp. 10\,933--10\,946.

\bibitem{jin2024one}
Y.~Jin, Z.~Cao, X.~Wang, J.~Wang, and M.~Long, ``One fits many: Class confusion loss for versatile domain adaptation,'' \emph{IEEE Transactions on Pattern Analysis and Machine Intelligence}, 2024.

\bibitem{wang2024unified}
Z.~Wang, Y.~Li, L.~Shen, and H.~Huang, ``A unified and general framework for continual learning,'' \emph{arXiv preprint arXiv:2403.13249}, 2024.

\bibitem{zhu2024model}
D.~Zhu, Z.~Sun, Z.~Li, T.~Shen, K.~Yan, S.~Ding, K.~Kuang, and C.~Wu, ``Model tailor: Mitigating catastrophic forgetting in multi-modal large language models,'' \emph{arXiv preprint arXiv:2402.12048}, 2024.

\bibitem{zhao2024sapt}
W.~Zhao, S.~Wang, Y.~Hu, Y.~Zhao \emph{et~al.}, ``Sapt: A shared attention framework for parameter-efficient continual learning of large language models,'' \emph{arXiv preprint arXiv:2401.08295}, 2024.

\bibitem{liang2024inflora}
Y.-S. Liang and W.-J. Li, ``Inflora: Interference-free low-rank adaptation for continual learning,'' \emph{arXiv preprint arXiv:2404.00228}, 2024.

\bibitem{razdaibiedina2022progressive}
A.~Razdai, Y.~Mao, R.~Hou, M.~Khabsa, M.~Lewis, and A.~Almahairi, ``Progressive prompts: Continual learning for language models,'' in \emph{The Eleventh International Conference on Learning Representations}, 2022.

\bibitem{wang2023orthogonal}
X.~Wang, T.~Chen, Q.~Ge, H.~Xia, R.~Bao, R.~Zheng, Q.~Zhang, T.~Gui, and X.~Huang, ``Orthogonal subspace learning for language model continual learning,'' \emph{arXiv preprint arXiv:2310.14152}, 2023.

\bibitem{feng2024tasl}
Y.~Feng, X.~Chu, Y.~Xu, G.~Shi, B.~Liu, and X.-M. Wu, ``Tasl: Continual dialog state tracking via task skill localization and consolidation,'' in \emph{Proceedings of the 62nd Annual Meeting of the Association for Computational Linguistics (Volume 1: Long Papers)}, 2024.

\bibitem{panigrahi2023task}
A.~Panigrahi, N.~Saunshi, H.~Zhao, and S.~Arora, ``Task-specific skill localization in fine-tuned language models,'' in \emph{International Conference on Machine Learning}.\hskip 1em plus 0.5em minus 0.4em\relax PMLR, 2023, pp. 27\,011--27\,033.

\bibitem{zhang2024unveiling}
Z.~Zhang, J.~Zhao, Q.~Zhang, T.~Gui, and X.~Huang, ``Unveiling linguistic regions in large language models,'' \emph{arXiv preprint arXiv:2402.14700}, 2024.

\bibitem{li2023crnet}
D.~Li and Z.~Zeng, ``Crnet: A fast continual learning framework with random theory,'' \emph{IEEE Transactions on Pattern Analysis and Machine Intelligence}, vol.~45, no.~9, pp. 10\,731--10\,744, 2023.

\bibitem{kirkpatrick2017overcoming}
J.~Kirkpatrick, R.~Pascanu, N.~Rabinowitz, J.~Veness, G.~Desjardins, A.~A. Rusu, K.~Milan, J.~Quan, T.~Ramalho, A.~Grabska-Barwinska \emph{et~al.}, ``Overcoming catastrophic forgetting in neural networks,'' \emph{Proceedings of the national academy of sciences}, 2017.

\bibitem{wang2024inscl}
Y.~Wang, Y.~Liu, C.~Shi, H.~Li, C.~Chen, H.~Lu, and Y.~Yang, ``Inscl: A data-efficient continual learning paradigm for fine-tuning large language models with instructions,'' \emph{arXiv preprint arXiv:2403.11435}, 2024.

\bibitem{pham2023continual}
Q.~Pham, C.~Liu, and S.~C. Hoi, ``Continual learning, fast and slow,'' \emph{IEEE Transactions on Pattern Analysis and Machine Intelligence}, 2023.

\bibitem{huang2024mitigating}
J.~Huang, L.~Cui, A.~Wang, C.~Yang, X.~Liao, L.~Song, J.~Yao, and J.~Su, ``Mitigating catastrophic forgetting in large language models with self-synthesized rehearsal,'' \emph{arXiv preprint arXiv:2403.01244}, 2024.

\bibitem{li2023variational}
X.~Li, S.~Wang, J.~Sun, and Z.~Xu, ``Variational data-free knowledge distillation for continual learning,'' \emph{IEEE Transactions on Pattern Analysis and Machine Intelligence}, vol.~45, no.~10, pp. 12\,618--12\,634, 2023.

\bibitem{rypesc2024divide}
G.~Rype{\'s}{\'c}, S.~Cygert, V.~Khan, T.~Trzci{\'n}ski, B.~Zieli{\'n}ski, and B.~Twardowski, ``Divide and not forget: Ensemble of selectively trained experts in continual learning,'' \emph{arXiv preprint arXiv:2401.10191}, 2024.

\bibitem{xu2021adaptive}
J.~Xu, J.~Ma, X.~Gao, and Z.~Zhu, ``Adaptive progressive continual learning,'' \emph{IEEE transactions on pattern analysis and machine intelligence}, vol.~44, no.~10, pp. 6715--6728, 2021.

\bibitem{chen2023towards}
C.~Chen, J.~Song, L.~Gao, and H.~T. Shen, ``Towards redundancy-free sub-networks in continual learning,'' \emph{arXiv preprint arXiv:2312.00840}, 2023.

\bibitem{li2024theory}
H.~Li, S.~Lin, L.~Duan, Y.~Liang, and N.~B. Shroff, ``Theory on mixture-of-experts in continual learning,'' \emph{arXiv preprint arXiv:2406.16437}.

\bibitem{michel2019sixteen}
P.~Michel, O.~Levy, and G.~Neubig, ``Are sixteen heads really better than one?'' \emph{Advances in neural information processing systems}, 2019.

\bibitem{chen2024learnable}
R.~Chen, T.~Hu, Y.~Feng, and Z.~Liu, ``Learnable privacy neurons localization in language models,'' \emph{arXiv preprint arXiv:2405.10989}.

\bibitem{aghajanyan2020intrinsic}
A.~Aghajanyan, L.~Zettlemoyer, and S.~Gupta, ``Intrinsic dimensionality explains the effectiveness of language model fine-tuning,'' \emph{arXiv preprint arXiv:2012.13255}, 2020.

\bibitem{qin2024large}
L.~Qin, Q.~Chen, X.~Feng, Y.~Wu, Y.~Zhang, Y.~Li, M.~Li, W.~Che, and P.~S. Yu, ``Large language models meet nlp: A survey,'' \emph{arXiv preprint arXiv:2405.12819}, 2024.

\bibitem{zhang2023adaptive}
Q.~Zhang, M.~Chen, A.~Bukharin, P.~He, Y.~Cheng, W.~Chen, and T.~Zhao, ``Adaptive budget allocation for parameter-efficient fine-tuning,'' in \emph{International Conference on Learning Representations}, 2023.

\bibitem{konishi2023spg}
T.~Konishi, M.~Kurokawa, C.~Ono, Z.~Ke, G.~Kim, and B.~Liu, ``{Parameter-Level Soft-Masking for Continual Learning},'' in \emph{Proc. of ICML}, 2023.

\bibitem{lecun1989optimal}
Y.~LeCun, J.~Denker, and S.~Solla, ``Optimal brain damage,'' \emph{Advances in neural information processing systems}, vol.~2, 1989.

\bibitem{rissanen1996fisher}
J.~J. Rissanen, ``Fisher information and stochastic complexity,'' \emph{IEEE transactions on information theory}, vol.~42, no.~1, pp. 40--47, 1996.

\bibitem{eddine2023weighted}
I.~Eddine~Marouf, S.~Roy, E.~Tartaglione, and S.~Lathuili{\`e}re, ``Weighted ensemble models are strong continual learners,'' \emph{arXiv e-prints}, pp. arXiv--2312, 2023.

\bibitem{wang2022super}
Y.~Wang, S.~Mishra, P.~Alipoormolabashi, Y.~Kordi, A.~Mirzaei, A.~Arunkumar, A.~Ashok, A.~S. Dhanasekaran, A.~Naik, D.~Stap \emph{et~al.}, ``Super-naturalinstructions: Generalization via declarative instructions on 1600+ nlp tasks,'' \emph{arXiv preprint arXiv:2204.07705}, 2022.

\bibitem{wang2022learning}
Z.~Wang, Z.~Zhang, C.-Y. Lee, H.~Zhang, R.~Sun, X.~Ren, G.~Su, V.~Perot, J.~Dy, and T.~Pfister, ``Learning to prompt for continual learning,'' in \emph{Proceedings of the IEEE/CVF Conference on Computer Vision and Pattern Recognition}, 2022, pp. 139--149.

\bibitem{qin2021lfpt5}
C.~Qin and S.~Joty, ``Lfpt5: A unified framework for lifelong few-shot language learning based on prompt tuning of t5,'' \emph{arXiv preprint arXiv:2110.07298}, 2021.

\bibitem{lin2004rouge}
C.-Y. Lin, ``Rouge: A package for automatic evaluation of summaries,'' in \emph{Text summarization branches out}, 2004, pp. 74--81.

\bibitem{lopez2017gradient}
D.~Lopez-Paz and M.~Ranzato, ``Gradient episodic memory for continual learning,'' \emph{Advances in neural information processing systems}, 2017.

\bibitem{ke2022continual}
Z.~Ke and B.~Liu, ``Continual learning of natural language processing tasks: A survey,'' \emph{arXiv preprint arXiv:2211.12701}, 2022.

\bibitem{farajtabar2020orthogonal}
M.~Farajtabar, N.~Azizan, A.~Mott, and A.~Li, ``Orthogonal gradient descent for continual learning,'' in \emph{International Conference on Artificial Intelligence and Statistics}.\hskip 1em plus 0.5em minus 0.4em\relax PMLR, 2020, pp. 3762--3773.

\bibitem{wang2024rehearsal}
M.~Wang, H.~Adel, L.~Lange, J.~Str{\"o}tgen, and H.~Sch{\"u}tze, ``Rehearsal-free modular and compositional continual learning for language models,'' \emph{arXiv preprint arXiv:2404.00790}, 2024.

\bibitem{raffel2020exploring}
C.~Raffel, N.~Shazeer, A.~Roberts, K.~Lee, S.~Narang, M.~Matena, Y.~Zhou, W.~Li, and P.~J. Liu, ``Exploring the limits of transfer learning with a unified text-to-text transformer,'' \emph{The Journal of Machine Learning Research}, vol.~21, no.~1, pp. 5485--5551, 2020.

\bibitem{touvron2023llama}
H.~Touvron, T.~Lavril, G.~Izacard, X.~Martinet, M.-A. Lachaux, T.~Lacroix, B.~Rozi{\`e}re, N.~Goyal, E.~Hambro, F.~Azhar \emph{et~al.}, ``Llama: Open and efficient foundation language models,'' \emph{arXiv preprint arXiv:2302.13971}, 2023.

\bibitem{sun2019lamol}
F.-K. Sun, C.-H. Ho, and H.-Y. Lee, ``Lamol: Language modeling for lifelong language learning,'' \emph{arXiv preprint arXiv:1909.03329}, 2019.

\bibitem{szegedy2016rethinking}
C.~Szegedy, V.~Vanhoucke, S.~Ioffe, J.~Shlens, and Z.~Wojna, ``Rethinking the inception architecture for computer vision,'' in \emph{Proceedings of the IEEE conference on computer vision and pattern recognition}, 2016.

\end{thebibliography}


\vspace{3pt}

\begin{IEEEbiography}[{\includegraphics[width=1in,height=1.25in,clip,keepaspectratio]{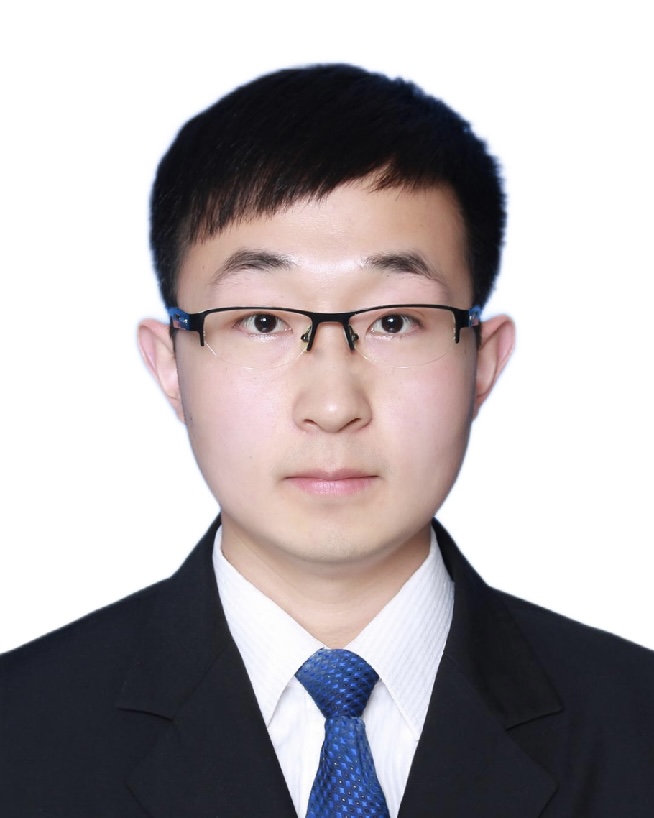}}]{Yujie Feng} received his master's degree from the School of Software and Microelectronics, Peking University, China, in 2022. He is currently a Ph.D. candidate at The Hong Kong Polytechnic University. His research interests focus on artificial intelligence and natural language processing.
\end{IEEEbiography}
 
\vspace{6pt}

\begin{IEEEbiography}[{\includegraphics[width=1in,height=1.25in,clip,keepaspectratio]{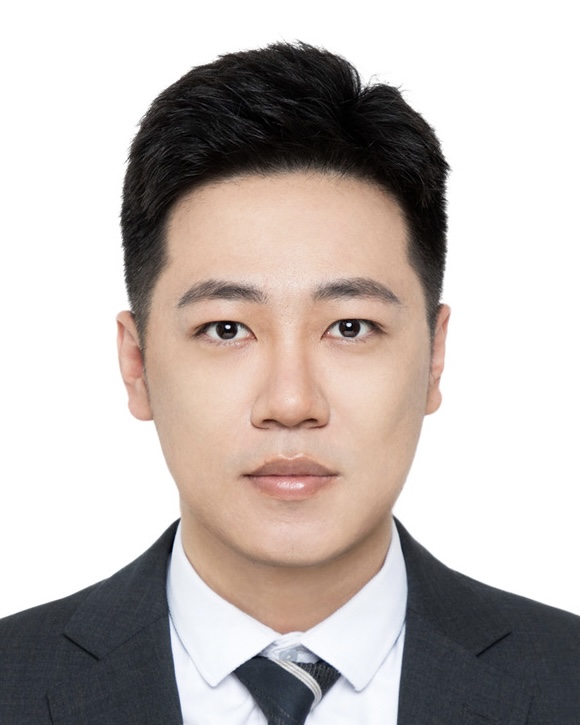}}]{Xu Chu} received the PhD degree from Peking University, Beijing, China. He is currently a Research Assistant Professor at the Frontier Center of the School of Computer Science, Peking University. His main research interests lie in the theory of machine learning and its algorithmic applications. He has published over 30 papers in top-tier conferences and journals, including ICML, NeurIPS, KDD, WWW, and TKDE, in the fields of artificial intelligence and data mining.
\end{IEEEbiography}
 
\vspace{3pt}

\begin{IEEEbiography}[{\includegraphics[width=1in,height=1.25in,clip,keepaspectratio]{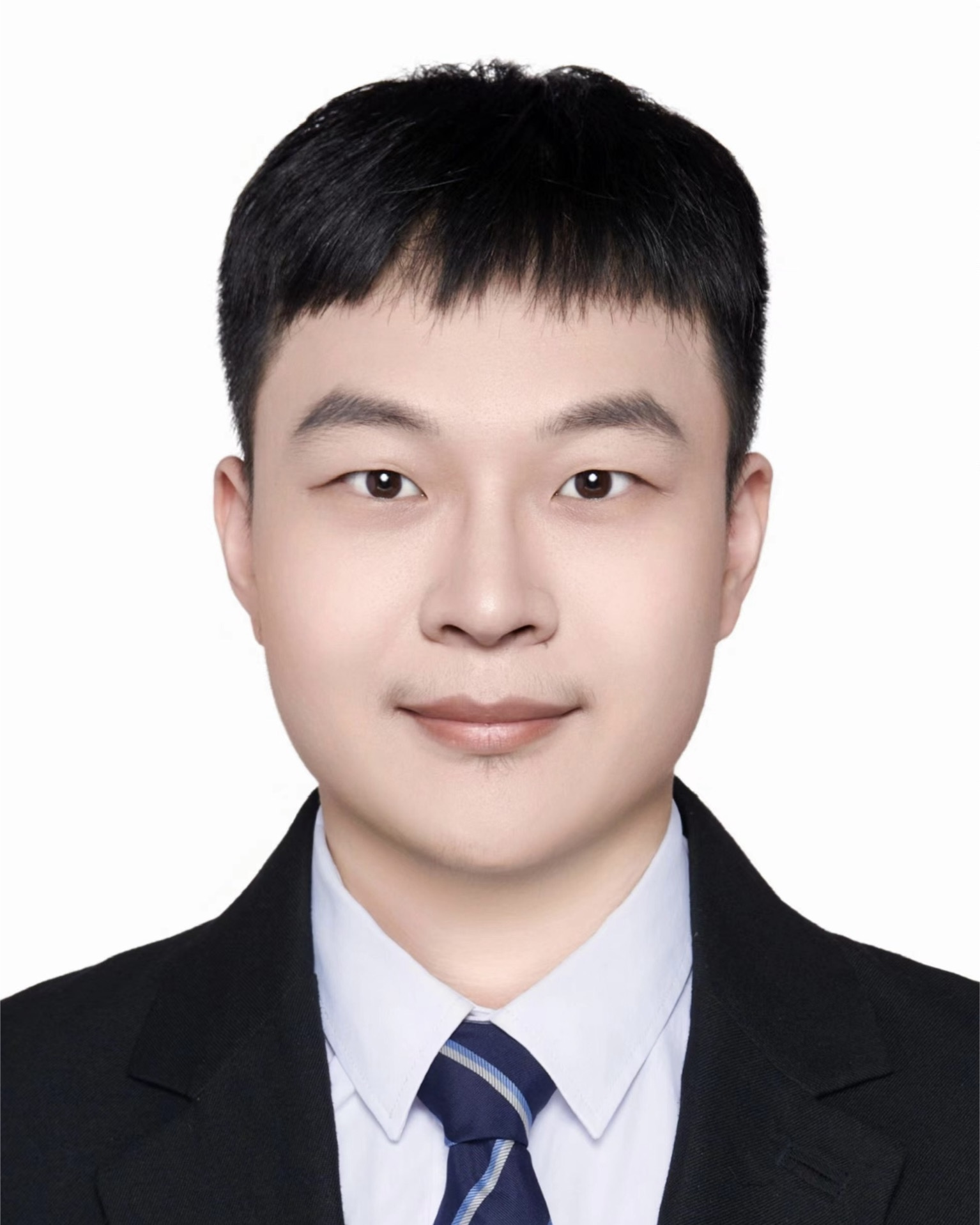}}]{Yongxin Xu} graduated from the School of Computer Science at Beijing University of Posts and Telecommunications in 2020 with a Bachelor's degree in Engineering. He is currently a PhD student at the School of Software and Microelectronics, Peking University. His research interests include large language models, knowledge graphs, and electronic medical record data analysis.
\end{IEEEbiography}
 
\vspace{20pt}

\begin{IEEEbiography}[{\includegraphics[width=1in,height=1.25in,clip,keepaspectratio]{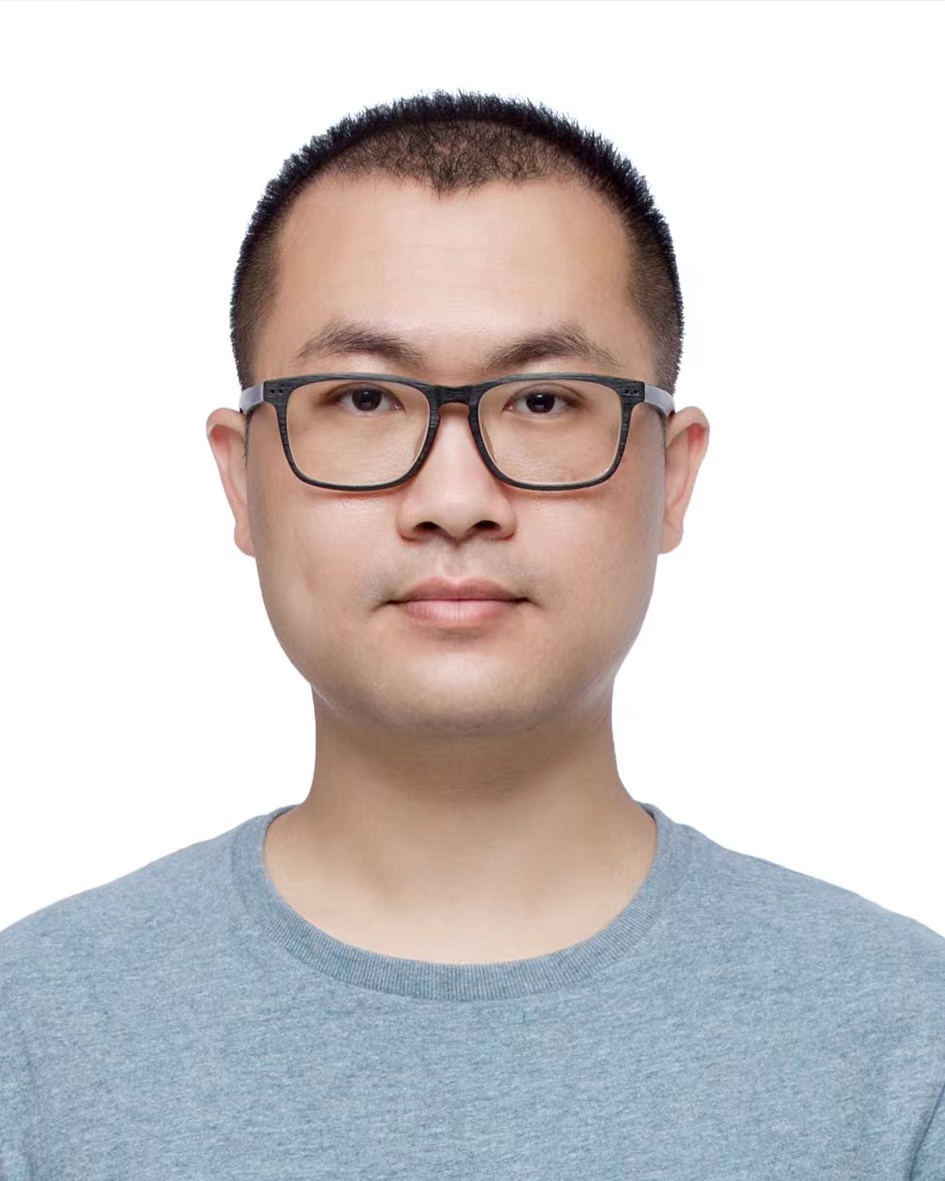}}]{Zexin Lu} received his PhD degree from the Department of Computing at The Hong Kong Polytechnic University in 2022. He is currently a postdoctoral researcher at the same university. He is the author or coauthor of papers presented at top conferences such as ACL, EMNLP, and COLING. His research interests focus on generative AI.
\end{IEEEbiography}

\vspace{20pt}
\begin{IEEEbiography}[{\includegraphics[width=1in,height=1.25in,clip,keepaspectratio]{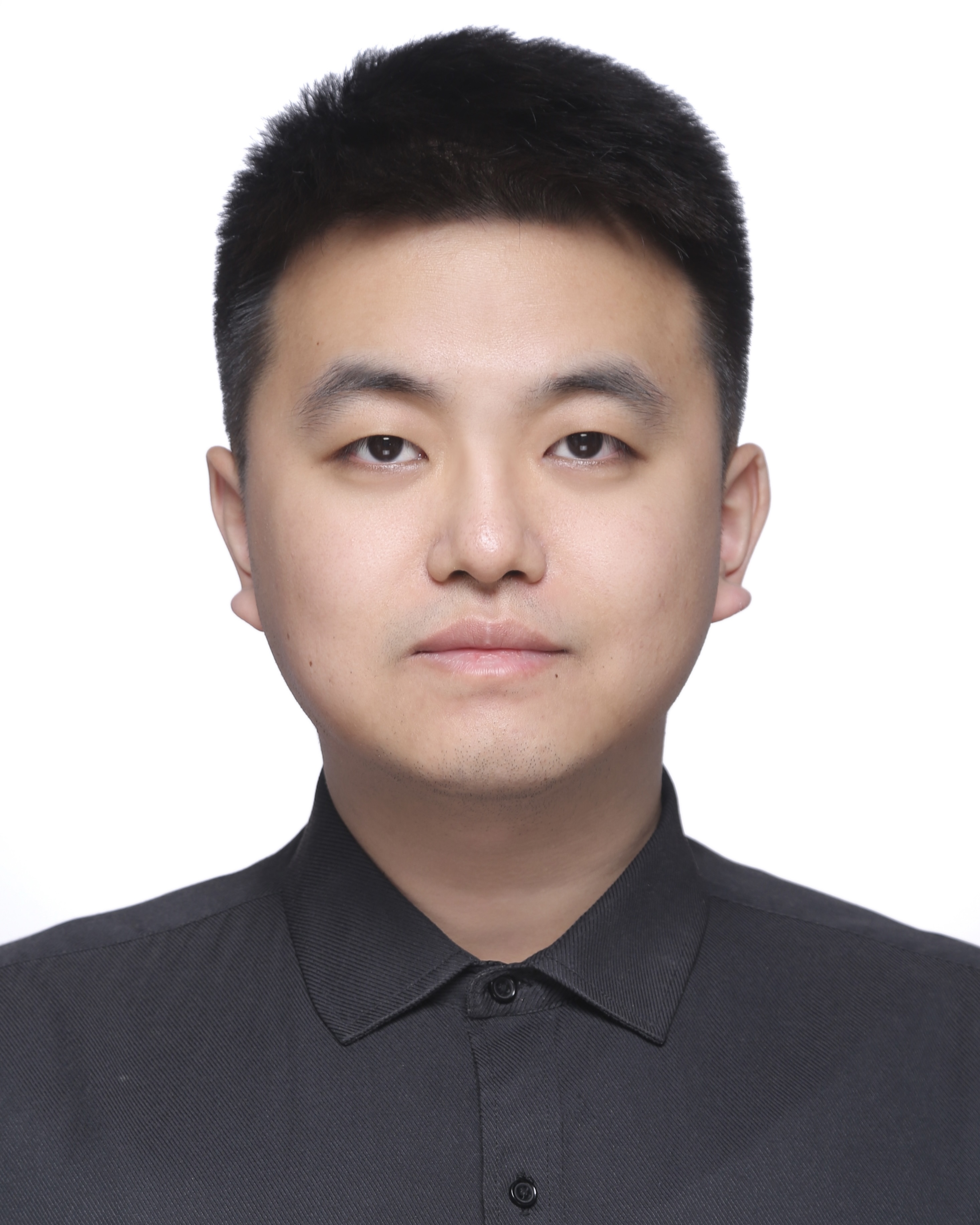}}]{Bo Liu} is now a Ph.D. candidate at Hong Kong Polytechnic Univerisity. He received his B.E. degree from Sichuan University in 2019. 
He is now focusing on multimodal learning and its application in healthcare.
\end{IEEEbiography}

\vspace{20pt}
\begin{IEEEbiography}[{\includegraphics[width=1in,height=1.25in,clip,keepaspectratio]{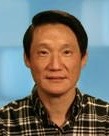}}]{Philip S. Yu} (Fellow, IEEE) received the PhD degree in electrical engineering from Stanford University. He is a Distinguished Professor in Computer Science at the University of Illinois at Chicago and also holds the Wexler Chair in Information Technology. His research interest is on big data, and artificial intelligence, including data mining, database and privacy. He is a Fellow of the ACM.
\end{IEEEbiography}

\vspace{20pt}

\begin{IEEEbiography}[{\includegraphics[width=1in,height=1.25in,clip,keepaspectratio]{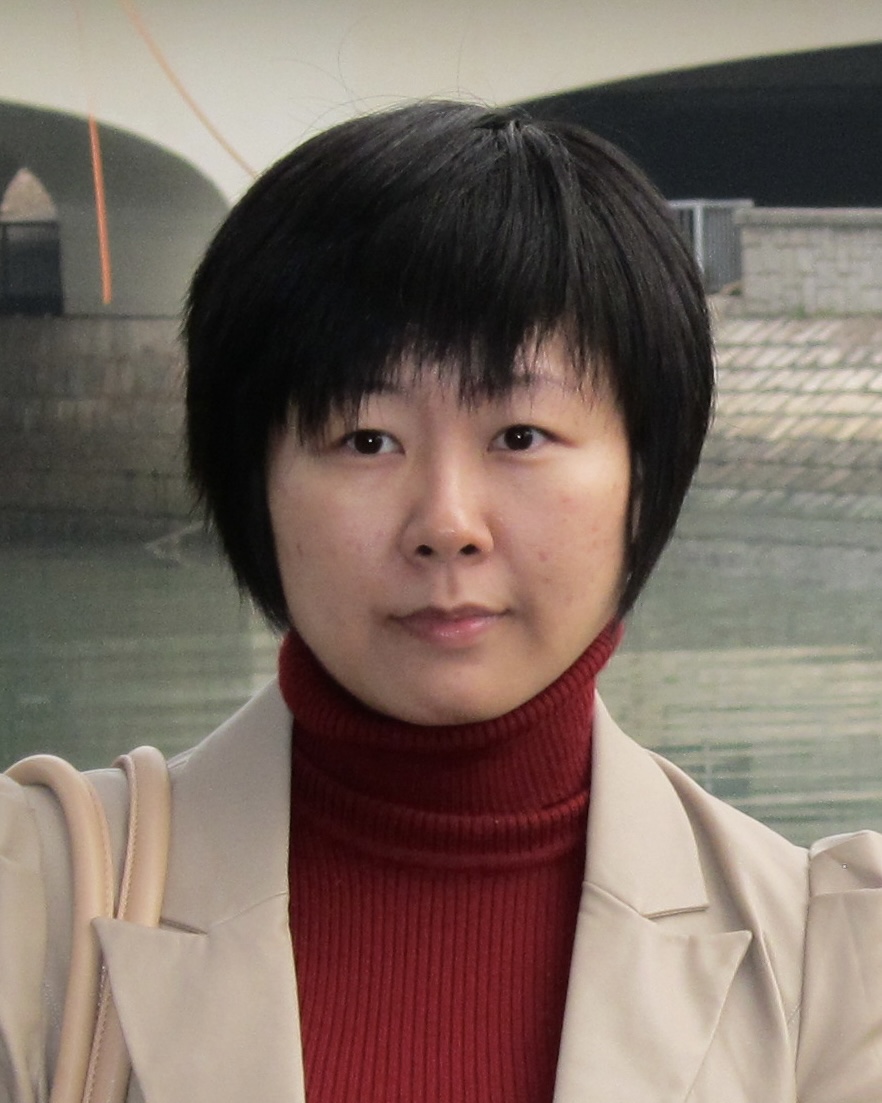}}]{Xiao-Ming Wu} (Senior Member, IEEE) is currently an Associate Professor at the Department of Data Science and Artificial Intelligence, The Hong Kong Polytechnic University. She has earned a PhD in Electrical Engineering from Columbia University in 2016, an MPhil from The Chinese University of Hong Kong, and holds both a BSc and an MSc from Peking University. 
Her recent research focuses on the development and applications of conversational, generative, and multimodal AI. 
\end{IEEEbiography}


\vfill

\end{document}